\documentclass[sigconf]{acmart}

\setcopyright{none} % No copyright notice required for submissions
\acmConference[Anonymous Submission to ACM CCS 2017]{ACM Conference on Computer and Communications Security}{Due 19 May 2017}{Dallas, Texas}
\acmYear{2017}

\usepackage{subcaption}

\begin{document}
%\title{Template for ACM CCS 2017} % TODO: replace with your title
\title{Adversarial Examples in Deep Learning: \\ Characterization and Divergence}
% \titlenote{Produces the permission block, and
% copyright information}
%\subtitle{Extended Abstract}
% \subtitlenote{The full version of the author's guide is available as
% \texttt{acmart.pdf} document}

% \author{Wenqi Wei}
% \author{Ling Liu}
% \author{Stacey Truex}
% \author{Lei Yu}
% \author{Mehmet Emre Gursoy}

% \author{Wenqi Wei \hspace{1cm} Ling Liu \hspace{1cm} Stacey Truex \hspace{1cm} Lei Yu \hspace{1cm} Mehmet Emre Gursoy \hspace{1cm} Yanzhao Wu}
% \affiliation{%
%   \institution{School of Computer Science, Georgia Institute of Technology}
%  % \streetaddress{P.O. Box 1212}
%   \city{Atlanta}
%   \state{GA}
%   \country{USA}
%   \postcode{30332}

\author{Wenqi Wei$^1$ \hspace{1cm} Ling Liu$^1$ \hspace{1cm} Margaret Loper$^2$ \hspace{1cm} Stacey Truex$^1$ \hspace{1cm} Lei Yu$^1$ \hspace{1cm} Mehmet Emre Gursoy$^1$ \hspace{1cm} Yanzhao Wu$^1$}
\affiliation{%
  \institution{$^1$School of Computer Science, Georgia Institute of Technology}
 % \streetaddress{P.O. Box 1212}
 \institution{$^2$Georgia Tech Research Institute(GTRI)}
  \city{Atlanta}
  \state{GA}
  \country{USA}
  \postcode{30332}
}

%\vspace{-14cm}

\begin{abstract}
The burgeoning success of deep learning has raised the security and privacy concerns as more and more tasks are accompanied with sensitive data. Adversarial attacks in deep learning have emerged as one of the dominating security threat to a range of mission-critical deep learning systems and applications. This paper takes a holistic and principled approach to perform statistical characterization of adversarial examples in deep learning. We provide a general formulation of adversarial examples and elaborate on the basic principle for adversarial attack algorithm design. We introduce easy and hard categorization of adversarial attacks to analyze the effectiveness of adversarial examples in terms of attack success rate, degree of change in adversarial perturbation, average entropy of prediction qualities, and fraction of adversarial examples that lead to successful attacks. We conduct extensive experimental study on adversarial behavior in easy and hard attacks under deep learning models with different hyperparameters and
different deep learning frameworks. We show that the same adversarial attack behaves differently under different hyperparameters and across different frameworks due to the different features learned under different deep learning model training process.
Our statistical characterization with strong empirical evidence provides a transformative enlightenment on mitigation strategies towards effective countermeasures against present and future adversarial attacks.

%% Deep learning has presented impressive results on a wide range of domains outperforming other machine learning approaches. However, recent work has shown that deep neural networks (DNNs) are vulnerable to adversarial examples, namely, artificial-perturbed inputs that are almost human-indistinguishable from the natural data and with the intent of being misclassifed by the DNN model. Such vulnerability seriously undermines the security assumption of deep learning powered system.
%%
%% In this paper, we perform a statistical and principled characterization on proposed attacks to help better understand them and to encourage defense method development. We provide a general formulation and basic principle for adversarial attack algorithm design, and statistically analyze the effectiveness of attack examples by dividing attack cases into hard and easy attacks. At last, we conduct extensive experiment study on adversarial behavior in easy and hard attacks under different hyperparameters. In brief, we provide a general view on adversarial behavior in deep learning to enlighten further research in understanding adversarial attacks and building defenses.

% Your abstract should go here. You will also need to upload a plain-text abstract into the web submission form.
\end{abstract}

% TODO: replace this section with code generated by the tool at https://dl.acm.org/ccs.cfm

% \ccsdesc[500]{Security and privacy~Usability in security and privacy}

% \ccsdesc{Security and privacy~Use https://dl.acm.org/ccs.cfm to generate actual concepts section for your paper}
% -- end of section to replace with generated code

 % TODO: replace with your keywords

\maketitle

\section{Introduction}

Deep learning has achieved impressive success on a wide range of domains like computer vision \cite{krizhevsky2012imagenet} and natural language processing \cite{collobert2008unified}, outperforming other machine learning approaches. Many of deep learning tasks, such as face recognition \cite{taigman2014deepface}, self-driving cars \cite{kim2017end}, speech recognition \cite{hinton2012deep} and malware detection \cite{dahl2013large}, are security-critical~\cite{barreno2006can, papernot2016towards}.

Recent studies have shown that deep learning models are vulnerable to adversarial input at prediction phase \cite{szegedy2013intriguing}. Adversarial examples are the input artifacts that are created from natural data by adding adversarial distortions. The purpose of adding such adversarial noise is to covertly fool deep learning model to misclassify the input.
For instance, attackers could use adversarial examples to confuse a face recognition authentication camera or a voice recognition system to breach a financial or government entity with misplaced authorization \cite{sharif2016accessorize}. Similarly, a self-driving vehicle could take unexpected action if the vision camera recognizes a stop sign, crafted by adversarial perturbation, as a speed limit sign, or if the voice instruction receiver misinterprets a compromised stop instruction as a drive-through instruction~\cite{carlini2016hidden}.

The threat of adversarial examples has inspired a sizable body of research on various attack algorithms \cite{madry2017towards, szegedy2013intriguing,kurakin2016adversarial,papernot2016limitations,evtimov2017robust,goodfellow2014explaining,carlini2017towards,sharif2016accessorize,carlini2016hidden,kurakin2016physical,xie2017adversarial,moosavi2016deepfool,elsayed2018adversarial,nguyen2015deep,ahmed2017poster}. Even with the black box access to the prediction API of a deep learning as a service, such as those provided by Amazon, Google or IBM, one could launch adversarial attacks to the privately trained deep neural network (DNN) model. Due to transferability  \cite{tramer2017space,liu2016delving,papernot2016transferability,papernot2017practical}, adversarial examples generated from one deep learning model can be transferred to fool other deep learning models. Given that deep learning is complex, there are many hidden spots that are not yet understood, such blind spots can be utilized as attack surfaces for generating adversarial examples. Furthermore,
adversarial attacks can happen in both training and prediction phases. Typical training phase attacks inject adversarial training data into the original training data to mis-train the network model \cite{huang2011adversarial}. Most of existing adversarial attacks are schemed at prediction phase, which is our focus in this paper.

%% Consequently,
To develop effective mitigation strategies against adversarial attacks, we articulate that the important first step is to gain in-depth understanding of the adverse effect and divergence of adversarial examples on the deep learning systems. In this paper, we take a holistic and principled approach to characterize adversarial attacks as an adversarial learning of the input data and a constrained optimization problem. We dive into the general formulations of adversarial examples and establish basic principles for adversarial noise injection. We characterize adversarial examples into easy and hard attacks based on statistical measures such as success rate, degree of change and prediction entropy, and analyze
different behavior of easy and hard cases under different hyperparameters, i.e., training epochs, sizes of feature maps and DNN frameworks. Moreover, we visualize the construction of adversarial examples and characterize their spatial and statistical features. We present empirical evidence on the effectiveness of adversarial attacks through extensive experiments. Our principled and statistical characterization of adversarial noise injection, the effectiveness of adversarial examples with easy and hard attack cases, and the impact of DNN on adversarial examples can be considered as enlightenment on the design of mitigation strategies and defense mechanisms against present and future adversarial attacks.

\section{Adversarial Examples and Attacks}
We first review the basic concept of DNN model and the threat model. Then we provide a general formulation of adversarial examples and attacks, describe the metrics for quantifying adversarial objectives and basic principle of adversarial perturbation.

\subsection{DNN Model}
Let $x \in X$ be an input example in the training dataset $X$. A DNN model $F(x)$ is made of $n$ successive layers of neurons from the input data to the output prediction, and $F(x) = f_n(\Omega_{n},f_{n-1} (\Omega_{n-1}, ... f_2 ( \Omega_2,\\f_1 (\Omega_1,x)))).$
Each layer represents a parametric function $f_i, i\in 1\dots n$, which computes an activation function, e.g., ReLU, of weighted representation of the input from previous layer to generate a new representation. The parameter set $\Omega_i, i\in 1\dots n$, consists of several weight vectors $\overrightarrow {W_i}=[\overrightarrow W_{k_i}]_{k_i\in1..K_i}$ and bias vectors $\overrightarrow {B_i}= [\overrightarrow B_{k_i}]_{k_i\in1..K_i}$, where $K_i$ denotes the number of neurons in layer $i$.
Let $Y=\{y^1, , ..., y^m\}$ be the class label space, where $m$ is the total number of labels, and $\overrightarrow y = F(x)$ be the classification prediction result for input $x$, in the form of m-dimension probability vector $\overrightarrow {y} = \{p_o, ..., p_j ..., p_m\}$, such that $p_j$ indicates the probability generated by DNN model toward class $y^j$, $0 \le {p_j} \le 1$ and $\sum\nolimits_{j = 0}^m {{p_j} = 1} $. The predicted label $C_{x}\in Y$ is the class with largest probability in the prediction vector. For ease of presentation, we assume that if there are multiple class labels with the same maximal probability, i.e., $\{y^s |s\in {1,\dots, m}, \forall j\neq s \in \{1,\dots, m\} s.t. p_{s} \ge p_j \}$, only one predicted class label is chosen for a given example $x$, denoted by $C_x$.
%% Note that it is possible that there are two or more labels that have the equally largest prediction probability. In this case, the predicted label is chosen either as the first among the largest or at random. However, we consider prediction label with unique largest probability only.
%%
The common output layer is a softmax layer, the prediction vector is also called logits, and we call the input of the softmax layer the prelogits.

%%Some deep learning tasks would study the top k(k>1) accuracy, which is more complex. But we only focus on situation where k=1.

The deployment of a DNN model consists of two phases: model training phase and model-based prediction phase. In training phase, a set of known input-label pairs $(x,y^L)$ is used to train a DNN model. $y^L$ is the known (ground truth) class label of the input $x$. The DNN first uses existing parameters $\Omega$ to generate classification from input data forwardly, then computes a loss function $J(\overrightarrow {y},y^L)$ that measures the distance between the prediction vector and the ground truth label.
With a goal of minimizing the loss function, the DNN training algorithm updates the parameters $\Omega$
at each layer using \emph{backpropagation} with an optimizer, e.g., stochastic gradient descent (SGD), in an iterative fashion.
The trained DNN will be refined through testing. Two metrics are used to measure the difference between the predicted vector and the ground truth label of the test input. One is accuracy that shows the percentage of test input whose predicted class ${C_x}\in Y$ is identical with its ground truth label $y^L$. The other is the loss function that computes the distance of the predicted class vector $\overrightarrow {y}$ to the label $y^L$. The trained DNN model produced at the end of the training phase will be used for classification prediction.
%% No space
At the prediction phase, the prediction API sends an input $x$ to the trained DNN model to compute its prediction vector $\overrightarrow {y}$ and the corresponding predicted label $C_{x}$ via the set of fixed parameters learned in the training phase.

 %% The prediction can be performed in either one-step or multi-step process. Compared to the one-step prediction, which may be too deterministic, multi-step prediction ensemble could improve the prediction accuracy by reducing speculation. For example, the user could randomly drop out some parameters from a small number of neurons at each one-step prediction, and produce a multi-step prediction ensemble by considering these one-step prediction together. Just like the dropout technique often used in the training process, the ensemble of the dropped-out prediction could generalize the result and mitigate the possible bias in one-step prediction.

\subsection{Threat Model}

\textbf{Insider Threat}: An insider threat refers to white-box attack and compromises from inside the organization that provides the DNN model based prediction service, such as poisoning attacks during training phase. Insider adversaries may know the DNN architecture, the intermediate results during the DNN computation, and are able to fully or partially manipulate the DNN model training process, e.g., injecting adversarial samples into the training dataset, manipulating the training outcome by controlling the inputs and outputs in some layers of DNN.
%%of the neurons or layers in DNN.

\textbf{Outsider Threat}: An outsider threat refers to black-box attack and compromises from external to a DNN model. Such attackers can only access the prediction API of the DNN as a service but do not have access to the DNN training and the trained DNN prediction model. However, attackers may have general and common background knowledge of DNN that is publicly available.
%% to perform query probing. In the context of adversarial examples in deep learning, we
We consider two types of outsider attacks: untargeted or targeted.

\textbf{Untargeted Attack} is a source class misclassification attack, which aims to misclassify the benign input by adding adversarial perturbation so that the predicted class of the benign input is changed to some other classes in $Y$ without a specific target class.
%% Namely, the adversary wants to generate an adversarial example by injecting adversarial perturbation to the benign input $x$ such that the predicted label of $x$ becomes any label other than the originally predicted class label of $x$.

\textbf{Targeted Attack} is a source-target misclassification, which aims to misclassify the predicted class of a benign example input $x$ to a targeted class in $Y$ by purposely crafting the benign input example $x$ via adversarial perturbation. As a result, the predicted class of the input $x$ is covertly changed from the original class (source) to the specific target class intended by the attacker.

\begin{figure}[ht]
  \centering
  \hspace*{-1cm}
    \includegraphics[scale=.55]{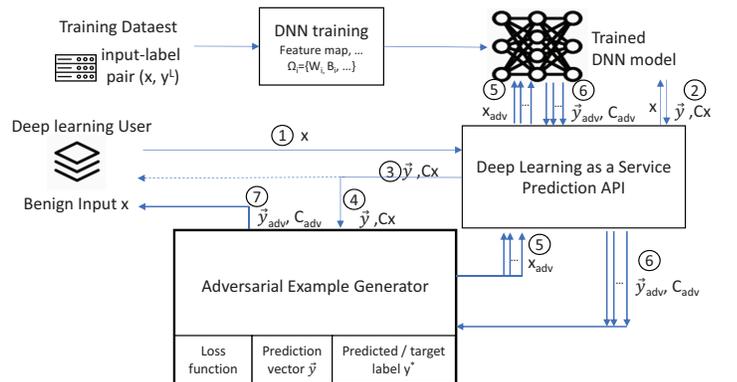}
    \vspace{-0.3cm}
    \caption{\small Outsider Adversarial Attack Workflow}
    \label{figure:attackworkflow}
    \vspace{-0.2cm}
\end{figure}

Let $x_{adv}$ be the maliciously crafted sample of benign input $x$. Figure~\ref{figure:attackworkflow} illustrates the workflow of an adversarial example based outsider attack to the prediction API of a deep learning service provider, consisting of 7 steps. (1) A benign input $x$ is sent to the prediction API, which (2) invokes the trained model to compute the prediction result. (3) Upon the API returning the prediction probability vector $\overrightarrow y$ and its predicted class label $C_x$, (4) the attacker may intercept the result and (5) launch an adversarial example based attack by generating adversarial example $x_{adv}$ (in one step or iteratively). (6) The adversary collects the prediction vector $\overrightarrow y_{adv}$ and the predicted class $C_{x_{adv}}$. The iteration stops if the attack is successful. (7) The user receives the incorrect result.
%\vspace{-0.5cm}
\vspace{-0.5cm}

\subsection{Formulation of Adversarial Examples}

For adversaries, an ideal adversarial attack is to construct a perturbed input $x_{adv}$ with minimal distance to $x$ to fool the DNN model such that the resulting prediction vector is ${\overrightarrow y}_{adv}$ with the predicted label $C_{x_{adv}}$, which is different from $C_x$, and yet its human user would still visually consider the adversarial input $x_{adv}$ similar to or the same as the benign input $x$ and thus believe that the predicted label $C_{adv}$ should be identical to $C_x$. Let $\Delta_x=dist(x,x_{adv})$ measure the distance between the original input $x$ and the adversarial input $x_{adv}$. Example distance metrics can be $L_0$, $L_2$, and $L_\infty$ norm. For image input, $L_0$ norm describes the number of pixels of $x$ that are changed; $L_2$ norm is the Euclidean distance between input $x$ and $x_{adv}$; and $L_\infty$ norm denotes the maximum change to any pixel of input $x$. $\Delta_x$ can be seen as the perturbation noise injected into the benign input to construct an adversarial input by the adversarial attack algorithm. We below provide a general formulation of an adversarial example based attack:
%\vspace{-0.3cm}
\begin{align}
&\quad \Delta_x=dist(x,x_{adv}) \label{equa: attack} \\
s.t. \quad &\min \,\beta\Delta_x + att(x_{adv})\,(1-\beta) g({\overrightarrow y}_{adv},y^{*}) \label{equa: attackconstraint} \\
& x \in X, \, x_{adv} \notin X \notag \\
&C_{x_{adv}} \ne C_{x}, \, C_{x} \in Y, \, C_{x_{adv}} \in Y/C_{x}  \notag\\
&HC_{x_{adv}} = HC_{x} \notag \\
\notag
\vspace{-0.4cm}
\end{align}
%\vspace{-0.3cm}

\noindent where $att(x_{adv})$ is a flag of untargeted and targeted attack: $att(x_{adv})$=-1 when the attack is untargeted and $att(x_{adv})$=1 when the attack is targeted. $ g({\overrightarrow y}_{adv},y^{*})$ is some objective function the attack seeks to optimize. Parameter $0 \le \beta \le 1$ controls the relative importance of the perturbation and the objective function. $HC_{x_{adv}} = HC_{x}$ means that human perceptional class of adversarial input needs to be the same as that of the original benign input.

For untageted attack, the label $y^{*}$ denotes the predicted class $C_{x}$ of the benign input $x$ such that $ g({\overrightarrow y}_{adv},C_{x})$ measures the distance between the prediction vector of adversarial input $x_{adv}$ and $C_x$. This distance is to be maximized to make the adversarial input successfully misclassified, thus achieving the goal of predicting the label $C_{x_{adv}}$ of adversarial input as any label but not the label of original benign input $x$. Accordingly, the perturbation $\Delta_x$ only concerns the prediction vector of adversarial example and the predicted class of benign example, defined as $\delta_x(\overrightarrow {y_{adv}},C_{x})$.

For targeted attack, the label $y^{*}$ denotes the target label $y^{T}$ so that $ g({\overrightarrow y}_{adv},y^T)$ is some objective function that measures the distance between the prediction vector of adversarial input $x_{adv}$ and the attack target label $y^{T}$. The attack is to minimize this distance so that the adversarial input $x_{adv}$, generated by maliciously injecting perturbation to $x$, is classified as the target label with high confidence. In contrast to untargeted attack, which only needs to lower the probability confidence on the original label $C_x$, the target attack needs to enhance the probability confidence on the target label while mitigating the probability confidence for all other labels so that the target label could stand out. This makes targeted attack much harder than untargeted ones. Thus, the adversarial noise $\Delta_x$ is generated from perturbation function $ \delta_x(\overrightarrow {y},C_{x},y^T)$.

Examples of objective function in Equation~\ref{equa: attackconstraint} include (1) $L_2$ loss, (2) cross-entropy loss, loss like (3) and their variants.

\vspace{-0.3cm}
\begin{align}
loss_{(1)}=&\sum \nolimits_j||p_j-q_j||_2, \quad p_j \in \overrightarrow y_{adv}, \, q_j \in \overrightarrow y^* \notag \\
loss_{(2)}=&-\sum \nolimits_j q_j \log p_j, \quad p_j \in \overrightarrow y_{adv}, \, q_j \in \overrightarrow y^*  \notag \\
loss_{(3)}=& \max(0, \max p_j-q^*), \quad p_j \in \overrightarrow y_{adv}, \, q^*=C_x \, \text{or} \, y^T \notag \\
\notag
\vspace{-0.4cm}
\end{align}

\vspace{-0.3cm}
\noindent where $\overrightarrow y^j$ is a one-hot prediction vector whose prediction class has the probability of 1 and the rest is 0, i.e., $p_j=1$, $\forall k \ne j, \, \: p_k=0$.

Equation~\ref{equa: attackconstraint} aims to optimize the objective function and the amount of perturbation at the same time while the parameter $\beta$ controls the relative importance of these two items.
When $\beta=1$, the problem is centered on minimizing the amount of perturbation while maintaining human imperceptibility.
Jacobian-based attack~\cite{papernot2016limitations} is an example.
When $\beta=0$, Equation~\ref{equa: attackconstraint} aims at optimizing the objective function without any regulation on the amount of perturbation.
It may suffer from over-crafting and lead to violation of human imperceptibility~\cite{sharif2016accessorize}.
When $0 < \beta < 1$, Equation~\ref{equa: attackconstraint} minimizes
the amount of perturbation and the objective function together to avoid either of them dominating the optimization problem.
Since large noise would increase the probability of attack being detected by human, the first term of Equation~\ref{equa: attackconstraint} acts as regularization of the second. In fact, the objective function could have an additional regularizer. Example attacks when $0 < \beta < 1$ are the Fast Gradient Sign Method (FGSM)~\cite{goodfellow2014explaining} and those optimization-based attacks~\cite{szegedy2013intriguing, carlini2017towards, evtimov2017robust}.

\subsection{Adverse Effect of Adversarial Examples}

We propose to use the following three evaluation metrics to analyze and compare the adverse effect of adversarial examples: Success Rate(SR), Degree of Change (DoC), and information entropy.

%%SR is the most popular metric in existing literature. It

SR measures the percentage of adversarial examples that result in successful attacks in all adversarial input data generated by using the adversarial example generator.
$$SR=\frac{\text{successful adversarial examples}}{\text{all adversarial examples}} $$
An adversarial example $x_{adv}$ is considered successful when under limited amount of perturbation noise, the following conditions hold: (1) $C_{adv}\neq C_x$; and (2) let $\overrightarrow y_{adv}=\{p_1,\dots, p_m\}$, for untargeted attack, $\exists k\in {1,\dots, m}$, we have $p_k \ge p_{C_x}$; and for targeted attack, $p_{y^T}$ is the maximum probability in the prediction vector $\overrightarrow y_{adv}$ such that $\forall j \in {1,\dots, m}$, $j\neq y^T$,  $p_{y^T} \ge p_j$ and $p_{y^T} > p_{C_x}$. By using SR, we are able to statistically measure how hard it is to move the original class of an input into another class. The higher the SR is, the more adverse effect the adversarial attack can cause. However, the SR alone is not sufficient to characterize the adverse effect of adversarial example. This motivates us to introduce per-class SR and other metrics.   Per-class SR measures the percentage of adversarial inputs that are successfully misclassified among the total adversarial examples generated from one class.

%% By setting 0.5 as the threshold of SR, it indicates that if more than 50\% of adversarial input are successfully misclassified, the attack is considered achieving a high SR. We measure class-wise SR, which means we feed adversarially crafted data from one class into deep learning model at a time and measure the percentage of inputs that are successfully misclassfied. However, the SR alone is not sufficient to characterize the adverse effect of adversarial example.

%%prediction vector exists a $p_*$ satisfying $p_* \ge p_{C}$ under limited amount of perturbation noise. This means $p_{C}$ is no longer the largest among all probabilities in $\{p_o, ..., p_m\}$ for untargeted attack, or $p_{C^T}$ becomes the maximum probability in the prediction vector $\overrightarrow y$ for targeted attack. A SR of 0.5 can be considered as threshold of high SR and low SR. That is, if more than 50\% of adversarial input are successfully misclassified, the attack is considered obtaining a high SR. The higher the success rate is, the more damage an adversary can cause. Though SR, we are able to statistically measure how hard it is to move the original class of an input into another class using success rate. This implies that success rate alone is not enough to characterize the effectiveness of adversarial attacks. We need to understand such effectiveness and its detrimental effect as well.

DoC describes the distance between adversarial example $x_{adv}$ and original benign input $x$, which is the objective that adversarial attacks aim to minimize in Equation \ref{equa: attackconstraint}. Let $\Delta _x = ||x_{adv}-x||_p$ be the $L_p$ distance of the two inputs, and the DoC is computed as
\begin{equation}
 DoC=\frac{1}{N}\sum\nolimits_1^N {\frac{\Delta_x}{||x||_p)}}, x_{adv}\in X_{adv}, |X_{adv}|=N
\end{equation}
where $||x||_p$ is the $p$ norm of the input data $x$. $p$ can be $0,2,\infty$. For example, if the perturbation noise changes 30 pixels in a 28*28-pixel image, the DoC for this perturbed image is $\frac{30}{28*28}=0.038$ under $L_0$ distance. If the dataset has 100 maliciously crafted images, i.e., $|X_{adv}|=N=100$, we add each of their DoC values and compute their average. From the perspective of adversaries, there are several advantages in keeping DoC theoretically the smallest. First, it means that less effort is required to launch the adversarial attack in a fast and efficient manner. Second, the minimal amount of change is one way to satisfy $HC_{x_{adv}} = HC_{x}$, making the attack human imperceptible and hard to detect. In contrast, randomly added perturbation can be inefficient for adversaries with high risk of making more effort adding noise with low SR.

%% And also, they might also get caught sabotaging the DNN model since the attack can be detected by human more easily.

The third metric is information entropy \cite{shannon1948mathematical}, defined as the average amount of information produced by a stochastic source of data. We compute the entropy on the distribution of probabilities in $\overrightarrow {y}$.
\begin{equation}
\Phi =  - \sum\nolimits_{j = 1}^m {{p_j}{{\log }_2}{p_j}},\quad p_j \in \overrightarrow {y}
\label{equa:entropy}
\end{equation}
The more even the probability is distributed, the larger the entropy is. Generally, entropy refers to disorder or uncertainty of a distribution. Thus the prediction vector of $[0.2,0.8,0,0,0,0,0,0,0,0]$ has larger entropy than that of $[0,1,0,0,0,0,0,0,0,0]$, because on the prediction of the second class, the latter is certain with a probability of 1 whereas the former is less certain with a probability of 0.8.
We also use the average information entropy, which is defined as follows: for $N$ input images, we compute their individual information entropy, accumulate the entropy of all $N$ inputs, and divide the entropy sum by $N$. The average entropy serves as a good alternative indicator to differentiate attacks with high and low SR.

%% Remove this para for CCS submission
%% For an adversarial example based attack to be successful, regardless it is untargeted or targeted, it needs to meet the following three minimum requirements: minimal amount of change, human imperceptibility, and high attack SR. These requirements are critical in characterizing adversarial examples, attacks and their divergence with respect to adverse effect.

\subsection{Basic Principle of Perturbation Injection}
\label{Section3.5}

%% We not only formally formulate adversarial attacks, we also interpret the adversarial attack process as gradient descent style adversarial perturbation update, which is an analogy to the training process of DNN. As mentioned, the input data can be seen as fixed in regular training, the parameters in the DNN model are updated with a goal to minimize the loss function and generate correct prediction. As counterpart in adversarial perturbing during prediction phase, the DNN parameters are fixed and the update is on the input data.

%% In Equation~\ref{equa: attack}, the goal of an adversarial example based attack
The goal of adversarial perturbation is to make DNN model to misclassify an input by changing the objective function value based on its gradients on the adversarial direction. Given the trained DNN has fixed parameters, the update should be on the benign input $x$ and be minimized so that only small amount of perturbation is added to the input $x$ to keep the human imperceptibility intact. Such adversarial perturbation update can be done in either one-step ($t=1$) or multiple steps iteratively ($t>1$):
%%, as formulated below:
\begin{equation}
x_{adv}^t =\left\{\begin{array}{l}
x+\theta R(\frac{\partial h(\overrightarrow y,y^*)}{{\partial {x}}}, \, t=1, \\
{x_{adv}}^{t - 1}+\theta R(\frac{\partial h({\overrightarrow y}_{adv}^{t-1},y^*)}{{\partial {x_{adv}}^{t - 1}}}), \, t> 1, \\
   \end{array} \right.
\label{equa: advupdate}
\end{equation}
For untargeted attacks, $y^*=C_{x}$. For targeted attacks, $y^*=y^T$. $\theta$ controls how much to update at a time.
%% , just like the learning rate in Equation~\ref{equa: update}.
$h(\,)$ is a function that describes the relation between the prediction vector ${\overrightarrow y}_{adv}$ of adversarial input $x_{adv}$ and some attack objective class $y^*$.
%% For example, $h(\,)$ can be the loss function, the prediction vector or some other objective function, like a loss function with a regularizer. Unlike $g(\,)$,
%%
$h(\,)$ concerns the prediction vector or loss function of the original input or the adversarial input in the previous iteration, which is the attack spot for adversarial perturbation update. In contrast, $g(\,)$  focuses on the prediction vector of the current adversarial input, which indicates the adversarial attack objective to be optimized. $R(\,)$ is the crafting rule that is based on partial gradient of function $h(\,)$, which implies how the perturbation is added. $R(\,)$ also ensures that the perturbation is clipped. Clip refers to a value constraint so that the perturbation cannot go beyond the range of the feature value. %%Consider image as an example,
For image, if the perturbation increases the value of a pixel beyond 255, then the pixel value is clipped to 255. The update term $\theta R(\,)$ has the same size as the input $x$ when $t=1$ and as the previous adversarial input $x_{adv}^{t - 1}$ when $t>1$.

{\bf One-step v.s. Multi-step Adversarial Examples.\/}
One-step attack is fast but excessive noise may be added to the benign input unnecessarily, because it is difficult to calibrate exactly how much noise is needed for successful attack.
%%By Equation~\ref{equa: attackconstraint}, o
One-step attack puts more weight on the objective function and less on minimizing the amount of perturbation. The partial gradient of $h({\overrightarrow y}_{adv}, y^*)$ in Equation \ref{equa: advupdate} only points out the direction of change and the relative importance of different places to perturb in the input data. Unlike one-step attack which has only one update on the original benign input, the multi-step attack uses a feedback-loop that iteratively modifies the input with more carefully injected perturbation until the input is successfully misclassified or the maximum perturbation threshold is reached to ensure the human imperceptibility ($H_{C_x}=H_{C_{adv}}$).
Although iterative attack is computationally more expensive, the attack is more strategic with high SR and less perturbation.
%%By Equation~\ref{equa: attackconstraint},
The multi-step attack strikes a good balance on minimizing the amount of perturbation and the objective function.
%% The decision on whether an attack is one-step or multi-step is a part of attack algorithm design for both untargeted attack and targeted attack.

%% We consider situations where $h(\,)$ is the loss function and the prediction vector, which are the mainstream in existing adversarial attack literature. Attack that works on other components of the DNN model could be an extension of our discussion.

{\bf Loss Function as the objective function.\/}
The loss function $J({\overrightarrow y},y^*)$
%%in Equation~\ref{equa: advupdate}
measures the distance between the prediction vector ${\overrightarrow y}$ and the attack destination class $y^*$. When $\frac{\partial J({\overrightarrow y},y^*)}{{\partial x}}=0$, it leads to minimal (local minimal) value of the convex (non-convex) loss function. When $\frac{\partial J({\overrightarrow y},y^*)}{{\partial x}}>0$ , adding values to input $x$ will increase the value of loss function. However, when $\frac{\partial J({\overrightarrow y},y^*)}{{\partial x}}<0$, adding values to input $x$ will decrease the value of loss function. Thus, manipulating the input $x$ could change the loss function.

\begin{figure}[ht]
\centerline{\includegraphics[width=7cm]{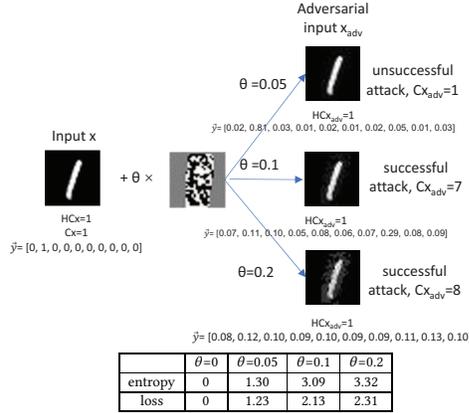}}
\vspace{0.05cm}
\scalebox{0.7}{
\small{
%\begin{tabular}{|c{1cm}|c{1cm}|c{1cm}|c{1cm}|c{1cm}|}
\begin{tabular}{|c|c|c|c|c|}
\hline
 & $\theta$=0 & $\theta$=0.05 & $\theta$=0.1 & $\theta$=0.2	 \\ \hline
entropy  & 0 & 1.30 & 3.09 &3.32 \\ \hline
loss & 0 & 1.23 & 2.13 & 2.31 \\ \hline
\end{tabular}
}
}
\vspace{-0.2cm}
\caption{{\small An illustration of loss function based attack ($\theta$=0.05, 0.1, 0.2), pixel values are added in light area of the perturbation, are deducted in dark area, and do not change in gray area.} }
\label{figure:FGSMonestepattack}
\vspace{-0.2cm}
\end{figure}

\begin{figure}[ht]
\centerline{\includegraphics[width=6.7cm]{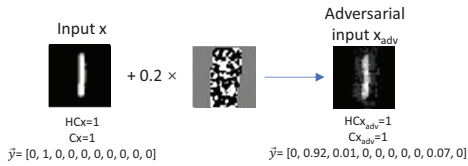}}
\vspace{-0.4cm}
\caption{{\small An unsuccessful loss function based attack ($\theta$=0.2)}}
\label{figure:FGSMunonestepattack}
\vspace{-0.2cm}
\end{figure}

Typical attacks that utilize the gradient of loss function include one step untargeted/targeted FGSM \cite{goodfellow2014explaining}, untargeted/targeted Iterative Method \cite{kurakin2016physical}, and the attack in \cite{sharif2016accessorize}.
To attack an image input by following \cite{goodfellow2014explaining}, we build a noise map based on the gradient of loss function and a simple pixel-based crafting rule $R(\,)$, in which the pixel value is set to 0 (dark) if the gradient of loss function at that pixel position is below zero, to 127 if the gradient is zero, and to 255 (light) if the gradient is above zero.
We are able to perform untargeted attacks by controlling the amount of noise injected using $\theta$.  Figure~\ref{figure:FGSMonestepattack} shows three adversarial examples (on the right) generated by applying the same perturbation noise (middle) to the same original input on the left under different $\theta$ values. When $\theta=0.05$, the attack fails. As we increase $\theta$ to $0.1$, the attack successfully misclassifies the input image of digit 1 to the class of digit 7, and
such misclassification is imperceptible as we visually still see the perturbed image on the right as digit 1 image without much change ($H_{C_x}=H_{C_{adv}}$).
By further increasing $\theta$ to $0.2$, the attack remains successful, but it misclassifies the same input image of digit 1 to the class of digit 8 instead. In Figure~\ref{figure:FGSMunonestepattack}, we use a different input image of digit 1, with the same $\theta=0.2$ and the same algorithm to add perturbation noise produced in the same way as in Figure~\ref{figure:FGSMonestepattack}. However, the attack is unsuccessful and the prediction vector indicates that the predicted class for the maliciously crafted example is still digit 1. This experiment shows that the success of attack is influenced both by $\theta$ and by the input instance. Furthermore, Figure \ref{figure:FGSMonestepattack} shows that both entropy and loss will grow with the larger gradient-based noise. Larger entropy indicates more even distribution of the probabilities in the prediction vector. Larger loss indicates that the prediction vector of adversarial input is more erroneous when compared with the predicted label of the benign input.
This empirical evidence also indicates that the entropy of the prediction vector and the loss is also a good metric for characterizing the effectiveness and divergence of attacks.

%\vspace{-0.3cm}
%\vspace{-0.4cm}

%\vspace{-0.2cm}
% \begin{figure}[ht]

% \begin{minipage}{0.2\linewidth}
%  \centerline{\includegraphics[width=1.2cm]{figure/original1.eps}}
%  \subcaption*{$\beta$=0}
% \end{minipage}
% \begin{minipage}{0.2\linewidth}
%  \centerline{\includegraphics[width=1.2cm]{figure/1theta005.eps}}
%   \subcaption*{$\beta$=0.05}
% \end{minipage}
% \begin{minipage}{0.2\linewidth}
%  \centerline{\includegraphics[width=1.2cm]{figure/1theta01.eps}}
%   \subcaption*{$\beta$=0.1}
% \end{minipage}
% \begin{minipage}{0.2\linewidth}
%  \centerline{\includegraphics[width=1.2cm]{figure/1theta02.eps}}
%   \subcaption*{$\beta$=0.2}
% \end{minipage}
% \caption{Loss function based noise injection, different $\theta$}
% \label{figure:FGSMoneimage}
% \end{figure}

% \begin{figure}[ht]
% \centerline{\includegraphics[width=9cm]{figure/differentnoiselevel.eps}}
% \vspace{-0.1cm}
% \caption{{\small Loss function based noise injection with varying $\theta$}}
% \label{figure:differentnoiselevel}
% \vspace{-0.2cm}
% \end{figure}

Optimization of loss function with regularization is an extension of the loss function based attacks using adversarial examples. CW attack \cite{carlini2017towards} and RP$_2$ attack \cite{evtimov2017robust} are examples of such attacks. The benefit of controlling the amount of change with a regularizer is to avoid drastic change. However, the optimization problem is often solved via techniques like Adam optimizer, which introduces considerable computation overhead.

{\bf Prediction vector based objective function.\/}
Similarly, for prediction vector, adding noise by increasing values on input $x$ will make the prediction vector value on class $j$ go up if the partial gradient of prediction vector on class $j$ is greater than zero, i.e., $\frac{\partial {y^j}}{{\partial {x}}}>0$. Analogously, $\frac{\partial {y^j}}{{\partial {x}}}<0$ means increasing the value of input will make the prediction vector value on class $j$ go down. When $\frac{\partial {y^j}}{{\partial {x}}}=0$, changing the input a little bit will not change the prediction vector at all. The same attack principle can be applied to prelogits as well, as prediction vector can be seen as normalized and difference-amplified prelogits.
%% Therefore, the adversaries can manipulate the prediction vector and prelogits.
Example attacks that utilize the gradient of prediction vector are
%%untargeted/targeted
Jacobian-based attack~\cite{papernot2016limitations} and Deepfool~\cite{moosavi2016deepfool}.
%%It is interesting to note that the
In fact, the gradient of loss function can be seen as an extension of gradient of prediction vector due to the chain rule since loss function is computed from prediction vector.
%%
%% Wenqi, we need to remove the entire summary and keep more experimental results in Section 4-5.

In summary, based on the inherent relation among input data, the gradient of loss function, and prediction vector, we are able to establish basic attack principle for untargeted and targeted adversarial attacks.
For {\em untargeted attack}, the goal is to craft the image in a way to lower the probability or confidence of the original class of the input until it is no longer the largest probability in the prediction vector. There are three ways to do so: (1) The adversaries could increase the loss function $J(\overrightarrow {y},y^{C_{x}})$, pushing the prediction $\overrightarrow {y}$ away from the predicted class $C_{x}$ of the benign input. When the loss function is large enough, the prediction will be changed to some destination class other than $C_{x}$. (2) One can decrease the value of $y^{C_{x}}$ in prelogits (feature vector) or logits (prediction vector), until this value is no longer the largest among all classes, the DNN prediction would misclassify the adversarial input. (3) By extending the loss function with a regularizer for the added noise, the objective function is optimized by increasing the loss function between the prediction vector of $x_{adv}$ and $y^{C_{x}}$, while minimizing the impact of perturbation.

Similarly, there are three methods for {\em targeted attack}: (1) The adversary may decrease the loss function $J(\overrightarrow {y},y^T)$ with perturbation, so that the crafting process is to perturb the input toward the target class $y^T$. (2) The adversary may increase the value of the prelogit or prediction vector of $y^T$ until it becomes the largest one, so that the DNN would misclassify the input into the target class. (3) The adversary extends loss function with optimization and decreases the loss function $J(\overrightarrow {y},y^T)$ while balancing the added noise.

%% Put the line spacing.

Table \ref{table:attacks} lists representative untargeted (U) and targeted (T) attacks with respect to noise origin, distance norm,  human perception (HP), and one-step or multi-step attack algorithm.

\vspace{-0.2cm}
\begin{table}[ht]
\centering
\scalebox{0.68}{
\small{
\begin{tabular}{|c|c|c|c|c|c|}
\hline
Attack &type &noise origin&distance norm&HP constrain & iteration	 \\ \hline
FGSM  & U/T & loss function & $L_\infty$ &$\theta$ & one\\ \hline
Iterative Method & U/T & loss function & $L_\infty$ & $\theta$ & multiple \\ \hline
Jacobian-based Attack & U/T & prelogit \& prediction & $L_0$& max iteration & multiple\\ \hline
CW \cite{carlini2017towards} & U/T & optimization & $L_0$, $L_2$, $L_\infty$& regularizer & multiple\\ \hline
RP$_2$ \cite{evtimov2017robust}  & U/T & optimization & $L_0$, $L_2$, $L_\infty$& regularizer & multiple\\ \hline
% FGSM  & untargeted/targeted & loss function & $L_2$ &$\theta$ \\ \hline
\end{tabular}
}}
\caption{\small Example Adversarial Attacks and Algorithms}
\label{table:attacks}
\vspace{-0.4cm}
\end{table}

\vspace{-0.4cm}

\section{One-Step Adversarial Examples}

%% In this section, we characterize one-step construction of adversarial examples and analyze their attack effectiveness and divergence.

% Using simple, grey-scale, and pixel-sparse MNIST dataset to study adversarial attacks could bring us more intuitive insight and motivate us to generalize the characterization to more complicated cases, e.g. color-rich, content-rich RGB images.

\subsection{One-Step Attack Generation}

We characterize one-step generation of adversarial examples on their attack effectiveness and divergence using FGSM. All experiments are conducted on MNIST dataset \cite{lecun2010mnist} with TensorFlow \cite{liu2018benchmarking}.
The adversarial update of FGSM is a special case of Equation \ref{equa: advupdate} with the crafting rule $R(\,)$ defined by function $sign(\,)$:

\vspace{-0.2cm}

\begin{align}
x_{adv} =& x+ \theta sign\,(\frac {\partial J(\overrightarrow {y},y^{C_{x}})}{\partial x}), \quad \text{untargeted} \notag \\
x_{adv} =& x- \theta sign\,(\frac {\partial J(\overrightarrow {y},y^{T})}{\partial x}), \quad \text{targeted} \notag
\vspace{-0.4cm}
\end{align}
In FGSM, $\theta$ serves as the DoC in $L_\infty$ distance and controls the amount of injected noise. Note that DoC in $L_0$ distance is different across instances as the gradient of loss function is determined by the input instance and the DNN model. The objective function $h(\,)$, which is the loss function, is to be optimized to achieve the attack goal.
The rule of perturbation noise injection depends on $sign(\,)$, which control the direction of the perturbation according to the objective of the loss function based attack.
For untargeted attack, pixel values should be decreased if $\frac {\partial J(\overrightarrow {y},y^{C_{x}})}{\partial x}$<0, and pixel values should be increased if $\frac {\partial J(\overrightarrow {y},y^{C_{x}})}{\partial x}$>0. Both are controlled by $sign(\,)$ and aim at increasing (maximizing) the loss function between the predicted vector and the predicted class label of the benign input $x$, which causes misclassification on $x_{adv}$.

\begin{figure*}
\centering
\includegraphics[scale=.61]{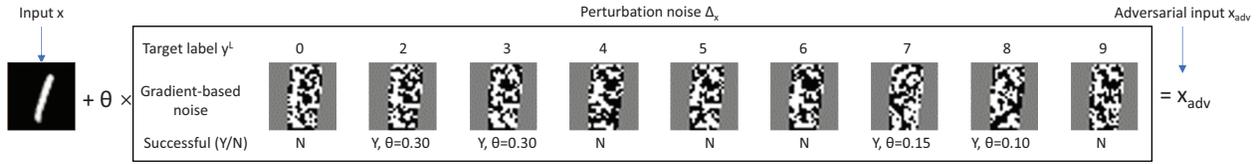}
\vspace{-0.4cm}
\caption{\small Visualization of Loss Function-Based Noise Injection for targeted FGSM attack}
\label{figure:signoneimage}
 \vspace{-0.4cm}
\end{figure*}

By slightly changing the crafting rule, targeted FGSM attacks can be established~\cite{kurakin2016physical}. The difference is that the loss function for targeted attack is defined between the prediction vector of a benign input $x$ and the target class of the attack. The direction of change is to decrease (minimize) the loss function so that the prediction moves towards the target class. We visualize the gradient of loss function for targeted FGSM attack in Figure \ref{figure:signoneimage}. It shows the pixel position whose value is to be increased when $\frac {\partial J(\overrightarrow {y},y^{C_{x}})}{\partial x}$<0 (dark area) and decreased when $\frac {\partial J(\overrightarrow {y},y^{C_{x}})}{\partial x}$>0 (light area).
Compared to untargeted attacks (recall Figure~\ref{figure:FGSMonestepattack} and Figure~\ref{figure:FGSMunonestepattack}),
Figure \ref{figure:signoneimage} shows that generating an adversarial example for successful targeted attack is much harder.
%%We conduct the set of experiments by varying $\theta$ from $0.1$ to $0.4$ with an interval of $0.05$
%%
When the targeted classes are set to digit 0, 4, 5, 6, and 9 respectively, we vary $\theta$ from $0.1$ to $0.4$ with an interval of $0.05$, the attacks are unsuccessful no matter how $\theta$ is set. When the targeted classes are set to digit 2, 3, 7, and 8 respectively, the attack succeeds at different noise level ($\theta$). The easiest target class with smallest noise level ($\theta=0.1$) is digit 8, followed by digit 7 ($\theta=0.15$), then digit 2 and digit 3 ($\theta=0.3$).
Larger $\theta$ beyond $0.4$ will cause clear violation of $H_{C_x}\neq H_{C_{adv}}$.

{\bf Takeaway Remarks.\/} The effectiveness of one-step attack heavily relies on $\theta$, which is the only parameter the adversary could fine-tune once the crafting rule $R(\,)$ is fixed.
%% It is critical for one-step attack to balance between the amount of noise added and the change of loss function value.
For untargeted attack, larger $\theta$ leads to larger update, and thus may result in over-crafting and violate the minimal amount of perturbation constraint. Also, large perturbation noise may dominate the original input,
%% such that the adversarial example is visually detectable by human,
%%
violating the human imperceptibility constraint. For targeted attack, only the correct amount of noise can lead to successful targeted misclassification. Over-crafting may cause the prediction go beyond the decision boundary of the target class, thus fail the attack. On the other hand, smaller $\theta$ leads to smaller perturbation, and higher chances of constructing adversarial examples that are not sufficiently perturbed to make attacks successful in one step.

As we have demonstrated in Figure~\ref{figure:FGSMonestepattack} and Figure~\ref{figure:FGSMunonestepattack}, $\theta$ is an important control parameter for the effectiveness of one step FGSM attack. For the same input, larger $\theta$ tends to increase SR. However, larger $\theta$ successful for attacking one input may not guarantee the success of attacking another input of the same class. Our analysis and characterization calibrate the effectiveness and divergence of adversarial examples with three interesting findings: (1) Different input images of the same class (e.g., digit 1) may have different attack effectiveness even with the same level of noise (same $\theta$) under the same attack method. This reflects the inherent problem of using fixed crating rules for all instances. The robustness against adversarial perturbation is different across input instances.
(2) For any benign input, there are more than one way to generate successful adversarial examples (e.g., using different $\theta$ values) using the same attack method.
(3) Different levels of noise (varying $\theta$) may lead to different attack effectiveness for the same input since the attack is not successful when $\theta=0.05$. Also, two different but successful attacks to the same benign input may lead to inconsistent misclassification results (recall Figure \ref{figure:FGSMonestepattack}, $C_{x}=1 \rightarrow C_{x_{adv}}=7 (\theta=0.1)$ v.s. $C_{x}=1 \rightarrow C_{x_{adv}}=8 (\theta=0.2)$). Moreover, different images from the same class can be misclassified into different destination classes (shown in Table \ref{table:untargeted10epoch}). It is critical to study such non-deterministic nature of adversarial examples for defense mechanism design.

\subsection{Characterization of One-step Attack}

We have shown that it is hard to balance between the amount of noise added and the change of loss function value in one step attack.
In this section, we further characterize the effectiveness and divergence of one-step adversarial examples by introducing a binary classification in terms of easy and hard attack categories for all successful attacks. This allows us to gain deeper understanding of both $\theta$ and the crafting rule $R(\,)$ with respect to SR, DoC, entropy, and the fraction of misclassified adversarial examples.
%% We argue that by differentiating different confidence levels in successful attacks, it will help us better characterize effectiveness and divergence of attacks and shed light on developing robust mitigation strategies.

We observe that the hardness for changing the predicted class of an input into another destination class varies for different destination classes under untargeted or targeted attacks.  We can distinguish successful attacks with higher SR as easy and efficient attacks and successful attacks with lower SR are hard and inefficient ones.
Table \ref{table:untargeted10epoch} shows a statistical characterization of easy and hard cases in one-step attacks with $\theta=0.2$.
%% The last column in each row gives the number of images perturbed under untargeted FGSM. For digits 1 and 2, they are $1135$ and $1032$ respectively.
The diagonal shows the percentage of adversarial examples that fail the attacks.
%%after the noise injection.
We consider two types of hardness in terms of different SRs: attack hardness based on source (S) classes and destination (D) classes.
For the attack hardness on source classes, it is observed that the SR of digit 1 is as high as 0.995, and the SR of digit 2 is as low as 0.771, indicating that $0.995*1135=1129$ images of digit 1 and $0.771*1032=796$ images of digit 2 succeed in source misclassification attack. We consider those attacks whose source SRs are relatively higher as one kind of easy cases. However, within each source class, different fractions of its $N$ images are misclassified into different destination classes, and such distribution is highly skewed for some cases. For source class of digit 1, the highest fraction that are misclassified into a particular destination class, i.e., digit 8, is 0.524 ($1135*0.524=595$ images), and the rest of the destination classes have significantly smaller fractions, ranging from 0.021 to 0.123. This indicates that the destination class of untargeted attacks is not uniformly random. We regard those successful attacks whose destination classes have higher fractions as another kind of easy cases.

\begin{figure}[ht]
  \centering
\includegraphics[scale=.25]{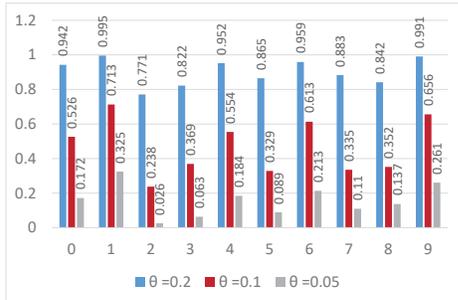}
\vspace{-0.4cm}
\caption{\small SR of untargeted FGSM with Different $\theta$: x-axis denotes the 10 classes and y-axis denotes SR.}
\label{figure: differenttheta}
 \vspace{-0.4cm}
\end{figure}

We first study the hardness in terms of source class. Although the per-class SR for each source class (e.g., digit 1) is different, Table \ref{table:untargeted10epoch} shows that one-step attack is highly effective and all SRs are relatively high with $\theta$=0.2. To better understand the attack hardness of source classes, we gradually lower the $\theta$ value from 0.2 to 0.1 and 0.05.
Figure~\ref{figure: differenttheta} shows the comparison of SR under different $\theta$. We highlight two observations. First, the SRs of all digits drop sharply as we decrease the $\theta$. Also when $\theta=0.05$, all source class attacks become hard as all SRs are smaller than 0.35.
Second, Digits 1 and 9 have higher SRs consistently with all three settings of $\theta$ compared to other digits. They are source class easy attacks under FGSM. We view this type of hardness as the vulnerability of the source class.
%% See \emph{Appendix A.1} for full sets of experiment with various $\theta$.

\begin{table}
\centering
\scalebox{0.65}{
\small{
\centering
\begin{tabular}{|c|c|c|c|c|c|c|c|c|c|c|c|c|}
    \hline
  S$\backslash$D & 0 &1 &2 &3 &4 &5 &6 &7 &8& 9 &SR & \# image \\ \hline
0 &0.058& 0.018& 0.067& 0.004& 0.015& \textbf{0.691}& 0.037& 0.027& 0.082& 0.001& 0.942 & 980\\ \hline
1 &0.068& 0.005& 0.123& 0.092& 0.015& 0.078& 0.006& 0.068& \textbf{0.524}& 0.021& 0.995 & 1135 \\ \hline
2 &0.010& 0.001& 0.229& 0.052& 0.049& \textbf{0.341}& 0.018& 0.065& 0.208& 0.027& 0.771 & 1032 \\ \hline
3& 0.032& 0.052& 0.192& 0.178& 0.022& \textbf{0.272}& 0& 0.031& 0.093& 0.128& 0.822 & 1010\\ \hline
4& 0.004 &0.017& 0.110& 0.063& 0.048& 0.045& 0.014& 0.049& \textbf{0.571}& 0.079& 0.952 & 982 \\ \hline
5 &0.047& 0.006& 0.142& 0.100& 0.074& 0.135& 0.029& 0.016& \textbf{0.341}& 0.110& 0.865 & 892\\ \hline
6 &0.056& 0.070& 0.217& 0.013& 0.116& 0.156& 0.041& 0.004& \textbf{0.307}& 0.02& 0.959 & 958\\ \hline
7 &0.031& 0.030& 0.314& 0.047& 0.018& 0.018& 0.001& 0.117& \textbf{0.385}& 0.039& 0.883 & 1028 \\ \hline
8 &0.020& 0.033&0.201& 0.031& 0.094& \textbf{0.380}& 0.015& 0.040& 0.158& 0.028& 0.842 & 974\\ \hline
9 &0.005& 0.016 &\textbf{0.336} &0.013 &0.046& 0.194 &0.003& 0.210& 0.168& 0.009& 0.991 & 1009 \\ \hline
\end{tabular}
}}
%\vspace{-0.1cm}
\caption{\small Untargeted FGSM Attack ($\theta$=0.2): the cell at $i^{th}$  row and $j^{th}$ column represents the fraction of adversarial inputs misclassifies source class in $i^{th}$ row to destination class in $j^{th}$ column.}
\label{table:untargeted10epoch}
 \vspace{-0.4cm}
\end{table}
%
%\vspace{-0.2cm}

%\vspace{-0.1cm}
%Combining the results from figure \ref{figure:FGSMonestepattack} and \ref{figure: differenttheta}, we can see that smaller degree of change for one-step attack will result in low SR while high degree of change could possibly make attack visible, just like the background noise shown in figure \ref{figure:FGSMonestepattack} when $\theta=0.2$. In fact, the perturbation of pixel value-sparse image is easy to discern and hard to attack. When performing FGSM on color-rich images, human perception on the noise is not as sensitive as the pixel-sparse ones, and people cannot perceive little perturbation.   Although this noise does not alter people's judgment on the input, people are aware of the existence of such tiny background perturbation noise that looks like colorful wireless channel noise on the analog television.
% (figure \ref{figure:channelnoise}).
% We encourage the readers to refer to attack examples in the NIPS 2017 Competition on Adversarial Attacks and Defences. In fact, human imperceptibility is a subtle concept depending on how much people can tolerate mutation.

% \begin{figure}[ht]
% \centerline{\includegraphics[width=7cm]{figure/channelnoise.eps}}
% \vspace{-0.1cm}
% \caption{{\small The Visibility of Perturbation ($\theta$=0.2)}}
% \label{figure:channelnoise}
% \vspace{-0.2cm}
% \end{figure}%%%this is the cat face image

%%
%%

The average information entropy is another statistical indicator to show the effectiveness of adversarial examples and the distribution of the prediction vectors. Table \ref{table:FGSMentropy} shows the average entropy of source classes under untargeted FGMS attack with different $\theta$ values. Clearly, the more vulnerable a source class is,  or the more successful the source misclassification attack is, the higher entropy is, showing more even distribution of the probabilities in the prediction vector(s).
Note that without attack, the entropy is as low as an order of $10^{-7}$. From both Figure~\ref{figure: differenttheta} and Table~\ref{table:FGSMentropy}, we can see that the most vulnerable source class is digit 1, followed by digits 9, 6, 4, 0, whose SRs are above 0.9. The next set of digits with SR above 0.8 is 8, 7, 5, 3, with digit 2 having the lowest SR of 0.773, which indicates that 234 inputs out of 1032 were failed under FGSM. We observe that FGSM attacks with high SRs result in more evenly distributed probabilities in prediction vector and larger entropy.
%% Similar to SR, some threshold can be set accordingly to differentiate easy and hard attacks but we have not found a universal threshold.
\vspace{-0.2cm}
\begin{table}[ht]
\centering
\scalebox{0.8}{
\small{
\begin{tabular}{|c|c|c|c|c|c|c|c|c|c|c|}
\hline
$\theta$ $\backslash$ S&	0	&1	&2	&3	&4	&5	&6	&7	&8	&9	 \\ \hline

0.05& 1.46& 1.96& 0.54& 1.47& 1.3& 1.06& 1.24& 0.93& 2.17 &2.26 \\ \hline
0.1& 2.72& 2.82& 1.37& 2.64& 2.54& 2.16& 2.3 &2& 3.02& 3\\ \hline
0.2 &2.97& 3.27& 2.44& 3.06& 3.12 &2.89& 2.96& 2.83 &3.11& 3.12 \\ \hline
\end{tabular}
}}
\caption{\small Entropy of FGSM under different $\theta$}
\label{table:FGSMentropy}
 \vspace{-0.4cm}
\end{table}
%%%%%%%%%%%this table can be in the form of figure.

\vspace{-0.2cm}

We next examine the second type of attack hardness with respect to destination (D) classes. Table \ref{table:top3untargeted} lists the top three easy and top three hard attacks. For source class digit 1, the top 3 easy destination classes are target digits 8, 2, and 3 with fraction of 0.524, 0.123 and 0.092 misclassified into digit 8 (1135*0.524=595 images), digit 2 (1135*0.123=139 images), and digit 3 (1135*0.092=104 images) respectively.
This indicates that the fraction of misclassified adversarial examples is a good indicator for characterization of the effectiveness of attacks.

%\vspace{-0.2cm}

\begin{table}[ht]
\centering
\scalebox{0.8}{
\small{
\begin{tabular}{|c|c|c|c|c|c|c|}
\hline
S &	Easy 1	& Easy 2 & Easy 3& Hard 1	&Hard 2	&Hard 3	 \\ \hline

0&	5/0.691&	8/0.082&	2/0.067&	9/0.001&	3/0.004&	4/0.015 \\ \hline
1&	8/0.524&	2/0.123&	3/0.092&	6/0.006&	4/0.015&	9/0.021 \\ \hline
2&	5/0.341&	8/0.208&	3/0.052&	1/0.001&	0/0.01&	6/0.016 \\ \hline
3&	5/0.272&	2/0.192&	9/0.128&	6/0.0&	4/0.025&	7/0.031\\ \hline
4&	8/0.571&	2/0.11&	9/0.079&	0/0.004&	6/0.014&	1/0.017 \\ \hline
5&	8/0.341&	2/0.142&	3,9/0.11&	1/0.006&	7/0.016&	6/0.029 \\ \hline
6&	8/0.307&	2/0.217&	5/0.156&	7/0.004&	3/0.013&	0/0.055 \\ \hline
7&	8/0.385&	2/0.314&	3/0.047	&6/0.001&	4/0.018&	5/0.018 \\ \hline
8&	5/0.38&	2/0.201&	4/0.094&	6/0.015&	9/0.028&	0/0.033 \\ \hline
9&	2/0.336&	7/0.21&	5/0.194&	6/0.003&	0/0.005&	3/0.013\\ \hline

\end{tabular}
}}
\caption{\small Top 3 Easy \& Hard Attacks under untargeted FGSM: each cell indicates the destination class digit and the fraction of adversarial examples being misclassified into that destination class.}
\label{table:top3untargeted}
 \vspace{-0.4cm}
\end{table}
%\vspace{-0.2cm}

\section{Multi-Step Adversarial Examples}
\label{multi-step}

In one-step generation of adversarial example, $\theta$ indicates to what extent an adversarial input is perturbed in one shot when the crafting rule is fixed. However, determining a right $\theta$ is non-trivial. Especially, smaller $\theta$ may lead to low SR or failure in one step attack. One remedy of achieving high attack SR with small $\theta$ is to use a multi-step boosting method.
%% (Section~\ref{Section3.5}).
The iteration process terminates when the misclassification goal is reached or when the perturbation violates $H_{C_{x}}=H_{C_{adv}}$.
The multi-step approach could increase SR significantly and make hard attacks at one-step easier.
Table~\ref{table:iterFGSM} shows the SR of multi-step iterative attack under untargeted FGMS with $\theta=0.005$, comparing the three settings of iterations.
%% (1 iteration is also in Figure \ref{figure: differenttheta}).
Since the loss function is computed at each iteration to fine-tune the attack direction, the noise injection is not simply repeating the previous iteration. With a few iterations, the attacker can significantly enhance SR. Hard cases in one-step attack may require more iterations to achieve certain SR goal comparable to one-step easy cases with high SR. Thus, the number of iterations can be a good indicator of the attack hardness.

%\vspace{-0.2cm}
\vspace{-0.3cm}
\begin{table}[ht]
\centering
\scalebox{0.78}{
\small{
\begin{tabular}{|c|c|c|c|c|c|c|c|c|c|c|}
\hline
iter $\backslash$ S&	0	&1	&2	&3	&4	&5	&6	&7	&8	&9	 \\ \hline
1& 0.172& 0.325& 0.026& 0.063& 0.184& 0.089& 0.213& 0.11& 0.137 &0.261 \\ \hline
3& 0.796& 0.921& 0.751& 0.789& 0.897 & 0.793& 0.903 &0.843& 0.775& 0.960\\ \hline
5 &0.988& 0.997& 0.924& 0.959& 0.971 &0.935& 0.982& 0.976 &0.937& 0.998 \\ \hline
\end{tabular}
}}
\caption{\small SR of Multi-step FGSM ($\theta=0.005$).}
\label{table:iterFGSM}
 %\vspace{-0.4cm}
\end{table}

\vspace{-0.3cm}
For targeted FGSM attacks, however, multi-step attack is less effective. The experiment conducted to attack input images of digit 1 fails to reach the target digit 0 for all 1135 images in MNIST. This is consistent with one-step targeted FGSM in Figure \ref{figure:signoneimage}, where one-step targeted attack is not successful even when $\theta$ reaches $0.4$. Consequently, boosting small $\theta$ iteratively may not improve SR of the attack when the attack under large $\theta$ is not successful. In addition to tuning $\theta$, the crafting rule may also need to be refined iteratively to boost attack SR.
%%, and larger $\theta$ value may cause violation of human imperceptibility. Hence,

%%not as powerful as it does in the untargeted attacks. In particular, all of the 1135 images of digit 1 fail to reach the targeted digit 0 in our experiment. We have shown that for all source class under untargeted attacks, the SR is low when $\theta$ is small and increases when $\theta$ goes up. However, the results in Figure \ref{figure:signoneimage} indicates that the attack is not successful even when the $\theta$ reaches 0.4. Consequently, the boosting of small $\theta$ iteratively does not improve the attack SR. In this case, tuning $\theta$ is not helpful and we might need to change the crafting rule to make attack successful.}}
%% Wenqi, why it is not true for targeted FGSM???
%% Iteratively performing an targeted attack is not as helpful as it does in untargeted attacks. Note that performing FGSM attack iteratively with a small DoC  will not change the relative distribution of easy and hard attacks, meaning that easy attacks are still easy even though the shares of portion are not exactly the same as they are in the one-step attack. In fact, easy attacks with more portion will have even more shares in iterative attacks. (Results provided in Appendix A.1.)

\subsection{Multi-Step Attack Generation}

 To better characterize the behavior of multi-step adversarial attack on its adverse effect and divergence, we further analyze the generation and effectiveness of multi-step adversarial example using targeted Jacobian-based attack ~\cite{papernot2016limitations}, which possesses a prediction vector-based objective function with two alternative crafting rules based on single pixel or a pair of pixels.

{\bf Single-pixel crafting rule.\/} Given input $x$ and its prediction vector $\overrightarrow {y}$, the attacker first computes the Jacobian matrix $Jac_F=\frac{\partial \overrightarrow y}{\partial x}=[\frac{\partial y^j}{\partial x}]_{j\in1 \dots m}$. Jacobian matrix on label $j$ indicates the relation between input features (pixels for image data) and the prediction on that label. That is, adding pixel value on one pixel would increase the value of the prediction $Y^j$ if the Jacobian matrix on class $j$ has a positive gradient on that pixel. Particularly, $\frac{\partial y^T}{\partial x},y^T \in \overrightarrow {y}$ is the Jacobian matrix for target class $T$. Since the prediction vector of the legitimate input is generated from the DNN model,  the gradient value is determined by the training process and the model is assumed to be differentiable. After computing the Jacobian matrix for the entire prediction vector, the adversary can compute the adversarial saliency map for target class $S(x,T)[\lambda]$ for each pixel $\lambda$.
\begin{equation}
%\vspace{-.1cm}
S(x,T)[\lambda] = \left\{ \begin{array}{l}
 0,\quad if \quad \frac{\partial y^T}{\partial x}[\lambda]<0 \quad or \quad \sum\nolimits_{j\not=T}\frac{\partial y^j}{\partial x}[\lambda]>0,\\
 \frac{\partial y^T}{\partial x}[\lambda]|\sum\nolimits_{j\not=T}\frac{\partial y^j}{\partial x}[\lambda]|,\quad otherwise,
 \end{array} \right.,
\label{equa:saliency}
\end{equation}
This equation gives four concrete gradient based crafting rules: (1) if adding pixel value does not move the prediction towards the target class, i.e., the gradient of prelogit or prediction for target class $p_T$ for the pixel is <0, or (2) the sum of all gradients other than that of the target class $p_j$ ($j \ne T$) for the pixel is >0, then the value on adversarial saliency map on that pixel is set to 0. However, if adding pixel value does move the prelogit and prediction towards the target class, i.e., (3) the gradient of logit of the target class for the pixel is >0, or (4) the sum of all gradients other than that of the target class for the pixel is <0, then the value of adversarial saliency map on that pixel is set to be the gradient product of (3) and (4). The power of adversarial saliency map is that it optimizes the objective function by considering both the gradient towards the target class and the gradient of all other classes.
Once the adversarial saliency map is computed, the adversary could craft the image with the pixels that have the largest adversarial saliency maps.

{\bf Pair-wise adversarial crafting rule.\/} The above adversarial saliency map considering individual pixels one at a time is too strict, especially when very few pixels would meet the heuristic search criteria in Equation \ref{equa:saliency}. Papernot, et al~\cite{papernot2016limitations} introduces the pair-wise adversarial saliency map. The heuristic searching for pairs of pixels is a greedy strategy that modifies one pair at a time. The incentive is the assumption that one pixel can compensate a minor flaw of the other pixel. For a pair of pixels $(\lambda_p,\lambda_q)$, we first compute its Jacobian matrices according to the prediction on each label. Then pair-wisely, we compute $\rm A$ and $\rm B$:
$$\rm A =\sum\nolimits_{i \in \{ p,q\} } {\frac{{\partial {y^T}}}{{{\lambda _i}}}}, \quad
\rm B =\sum\nolimits_{i \in \{ p,q\} } {\sum\nolimits_{j \ne T} {\frac{{\partial {y^j}}}{{{\lambda _i}}}} },$$
where  $\rm A$ represents to what extent changing these two pixels will change the prediction on the target class. $\rm B$ denotes the impact of changing the two pixels on classes other than the target.
Similar to adversarial saliency map, pixel pair with the largest value on $- \rm A \times \rm B$ when $\rm A>0 $ and $\rm B<0$ is chosen to be crafted.
%% No spacing.
The perturbation sets the pixel value to 255. A dynamic search domain is maintained to keep track of those pixels whose values already reach 255. The multi-step perturbation process iterates until the attack is successful or reaches the pre-defined maximum level of noise tolerance, such as 15\% of pixels. For image of $28\times 28$,  the adversarial example changes up to 28*28*0.15=118 pixels. If the adversarial input reaches this maximum noise level but is still not predicted as the target label, the attack fails.

\begin{figure*}
\centering
\includegraphics[scale=.66]{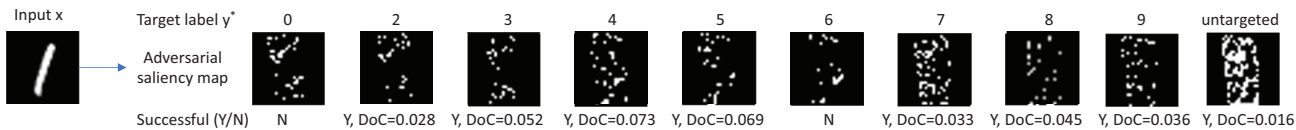}
\vspace{-0.4cm}
\caption{\small Visualization of Adversarial Saliency Map-based Noise Injection for targeted attacks. The Adversarial Saliency Map shown is from the first iteration. The noise of digit 1 is for untargted attack.}
\label{figure:saliencyoneimage}
 \vspace{-0.4cm}
\end{figure*}

Figure \ref{figure:saliencyoneimage} visualizes the adversarial saliency map for targeted attacks with each of the classes other than the source digit 1 as the target in the first iteration. Moreover, the adversarial saliency map of untargeted Jacbian-based attack is provided for reference. The construction is straightforward according to the general principle of perturbation injection. Since the value of adversarial saliency map is widely ranged, it is hard to demonstrate its numerical difference. For better visualization, we set all pixels whose values are non-zero in the saliency map towards the target class to 255. This means pixels in light area are potential candidates
for crafting. The visualization shows clearly that the saliency map perturbation is different across the attack target class, and the DoC values for successful targeted attacks are different. Target attacks to digits 0 and 6 are failed with the 15\% pixel-level perturbation threshold.

\begin{figure}[ht]
 \centerline{\includegraphics[width=6.2cm]{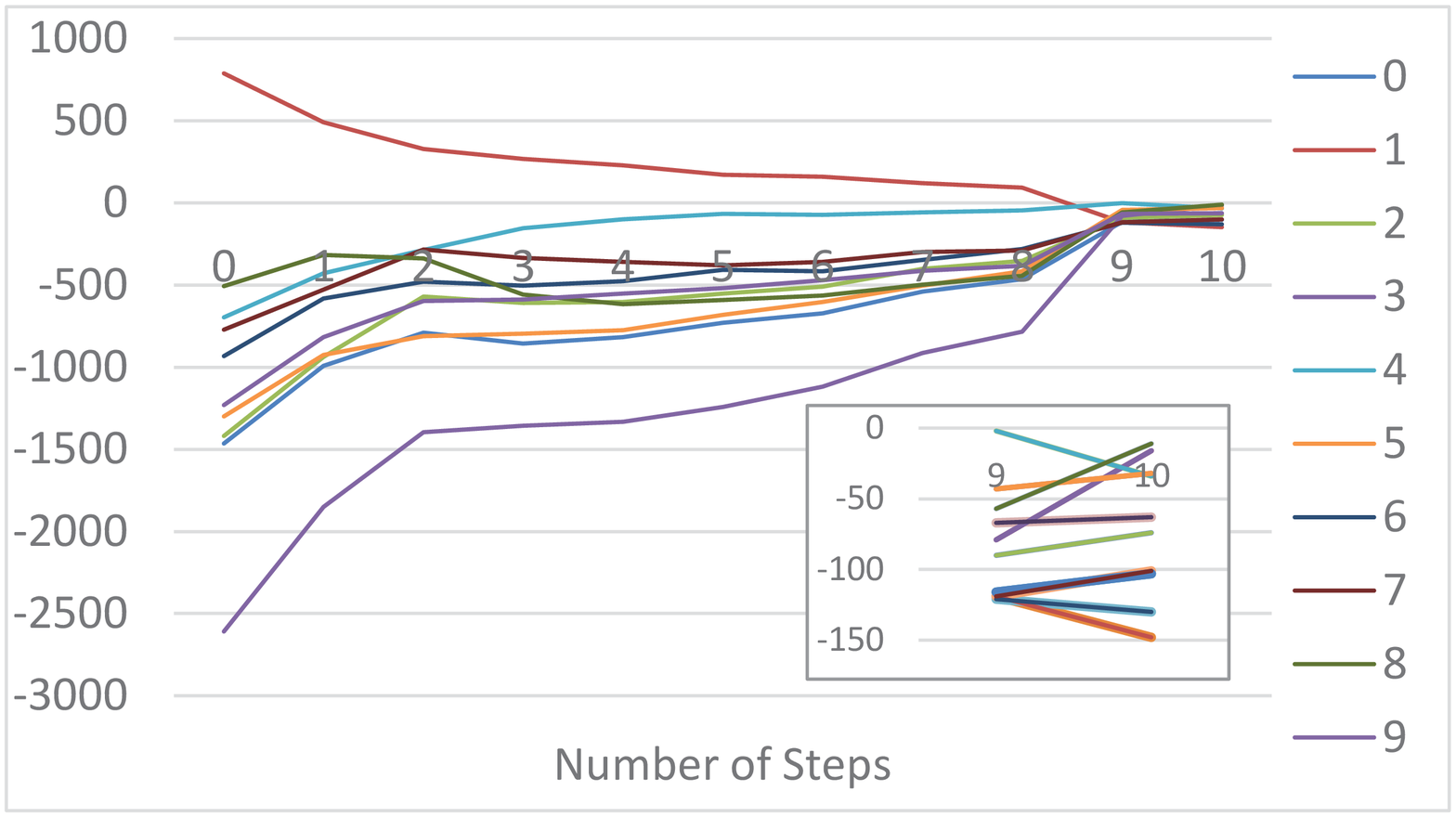}}
 \vspace{-0.1cm}
 \caption{\small Prelogits over 10 steps (source digit 1, target digit 8)}
\label{figure: JAMAoneimageprelogit}
 \end{figure}

 \vspace{-0.3cm}
 \begin{table}[ht]
\centering
\scalebox{0.75}{
\small{
\begin{tabular}{|c|c|c|c|c|c|c|c|c|c|c|c|}
    \hline
   $y^*$ $\backslash$ step&	0&	1&	2&	3&	4&	5&	6&	7&	8&	9 & 10 \\ \hline
  1 & 788 & 490 & 329& 268 & 228 & 171  & 160 & 119 & 94 & -119 & -148 \\ \hline
 4& -697 & -430 & -288 & -156 & -99 & -66 & -74 & -58 & -47 & -2 & -34 \\ \hline
 8 & -507 & -318 & -339 & -557 & -617 & -591 & -564 & -497 & -443 & -57 & -11 \\ \hline
\end{tabular}
}}
%\vspace{-0.4cm}
\caption{\small Prelogits trajectory of three representative classes (digits 1, 4, 8) over 10 iteration steps (source digit 1, target digit 8)}
\label{table: JAMAoneimageprelogit}
%\vspace{-0.4cm}
\end{table}

\vspace{-0.3cm}

We use a successful 10-step Jacobian-based targeted attack from source class 1 to target class 8 as an example to characterize the multi-step Jacobian noise injection.
Table~\ref{table: JAMAoneimageprelogit} shows three representative classes over the 10-step targeted attack and Figure~\ref{figure: JAMAoneimageprelogit} shows the prelogit trajectory for all classes in the 10-step attack.
At each step, an adversarial saliency map is computed and the pixel position with largest value is chosen to be the crafting pixel. It is a nested multi-step ensemble process with every step correcting the perturbation trajectory path a little bit. Though the perturbation is based on the gradient of prediction vector, the prediction vector remains to be  [0,1,0,0,0,0,0,0,0,0] for the first 8 steps, and changes to  [0,0,0,0,1,0,0,0,0,0] (predicted as digit 4) at the 9th step, and to [0, 0, 0, 0.0067, 0, 0, 0, 0, 0.9933, 0] at the 10th step, successfully reaching the digit 8 goal of this targeted attack.
It is observed that the trajectory of prelogits is much smoother and more informative compared to the drastic change in prediction vector in the iterative ensemble learning process. According to  Figure~\ref{figure: JAMAoneimageprelogit}, the perturbation enhances the prediction of targeted class while the gap between prelogits for different labels shrinks as the attack progresses step by step to the $10^{th}$ iteration. The prelogit of target class 8 gradually becomes the largest, succeeding in misclassifying digit 1 to digit 8 at the $10^{th}$ step.

%% Wenqi, let‚Äôs remove all appendixes due to timing to better organize them.
%% We provide Table \ref{table: JSMAoneimagelogit} in Appendix~\ref{JSMAprocess} to show how the prediction vector changes in all classes as the iterative attack progresses. It shows that although the prelogits in Table \ref{table: JAMAoneimageprelogit} changes drastically, the change of prediction vector is small and more concentrated but effective for the attack.

{\bf Takeaway Remarks.\/}
%%This in-depth analysis of Jacobian multi-step targeted attack provides us deeper insight on the effectiveness characterization and divergence of multi-step adversarial attacks.
(1) The pixel-level perturbation changes the prelogits, the prediction vector, and the probability of every single class in each step. This increases the hardness of targeted attack when the number of classes is large, as it is more difficult to choose the fraction of pixels as the maximum crafting cap, and very likely 15\% of pixel perturbation may lead to low SR or failure of the attack. (2) While softmax layer amplifies the numerical difference in prelogits and normalizes them into the prediction vectors, performing prediction vector based attack is equivalent to performing attacks via prelogits.
More interestingly, by noticing that the gap of prelogits converges for several successive prediction attempts, it may reveal the presence of a targeted attack.
(3) The pixel with larger adversarial saliency map indicates that adding its value will move the prelogit or prediction toward the target class more. This crafting rule, together with a limitation on the maximum iteration ensures minimal amount of perturbation as well as human imperceptibility. However, it also exposes some problems of this attack. In addition to its computation inefficiency, the perturbation simply sets the chosen pixel values to 255 at each iteration, which may over craft the input at times so that the noise deviates too much from the amount needed and results in unsuccessful attacks. Also adding full 255 to a pixel each time may not be effective or human-imperceptible for colored images.

\subsection{Effectiveness of Multi-step Attack}

In this section we characterize the effectiveness and divergence of multi-step targeted attack by analyzing the easy and hard cases.
%% using the Jacobian-based method for illustration.

\begin{figure}[ht]
 \begin{minipage}{0.50\linewidth}
  \centering
    \includegraphics[scale=.17]{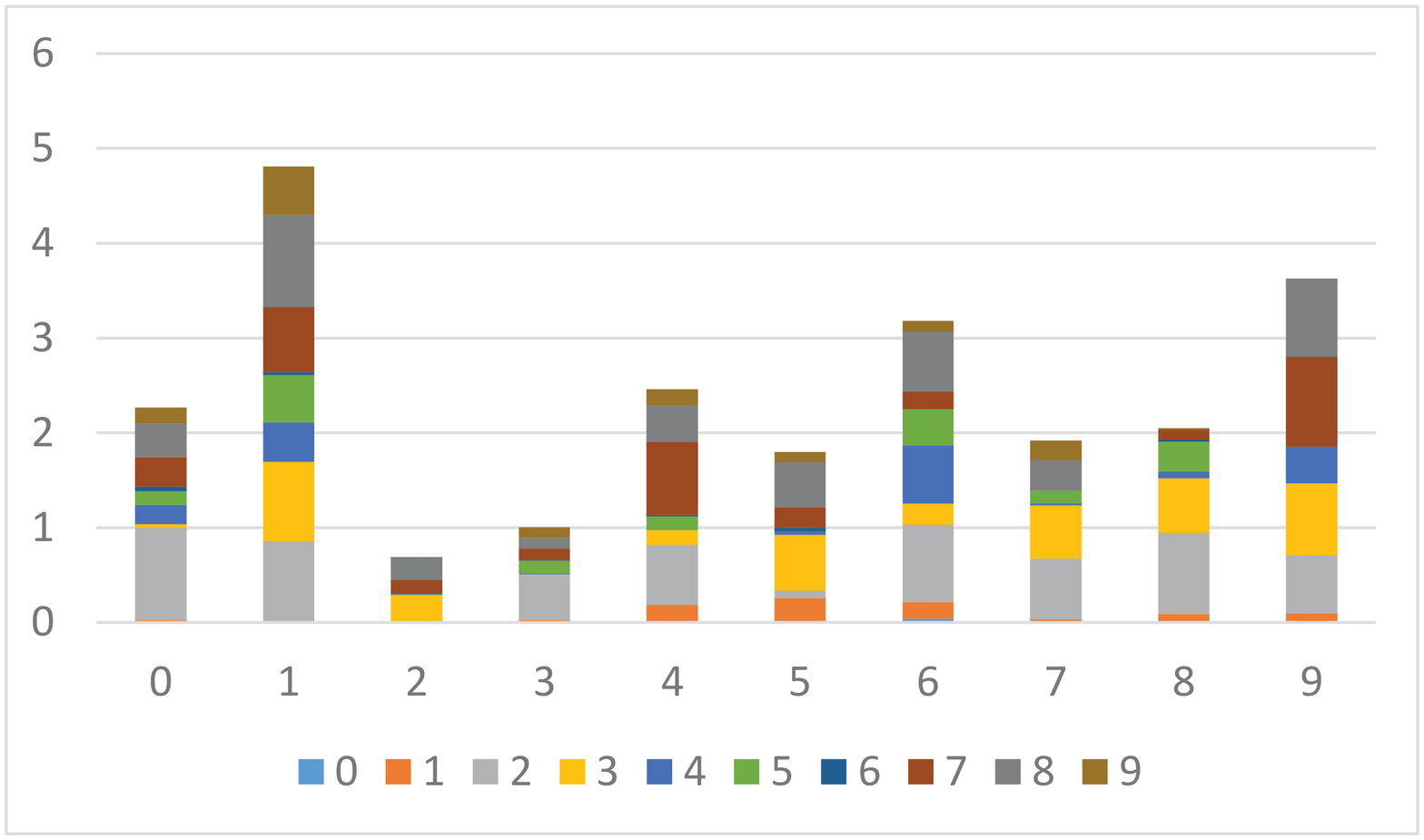}
    \vspace{-0.2cm}
    \caption{\small Source Vulnerability}
    \label{figure: 10eposource}
 \end{minipage}
 \begin{minipage}{0.49\linewidth}
  \centering
    \includegraphics[scale=.16]{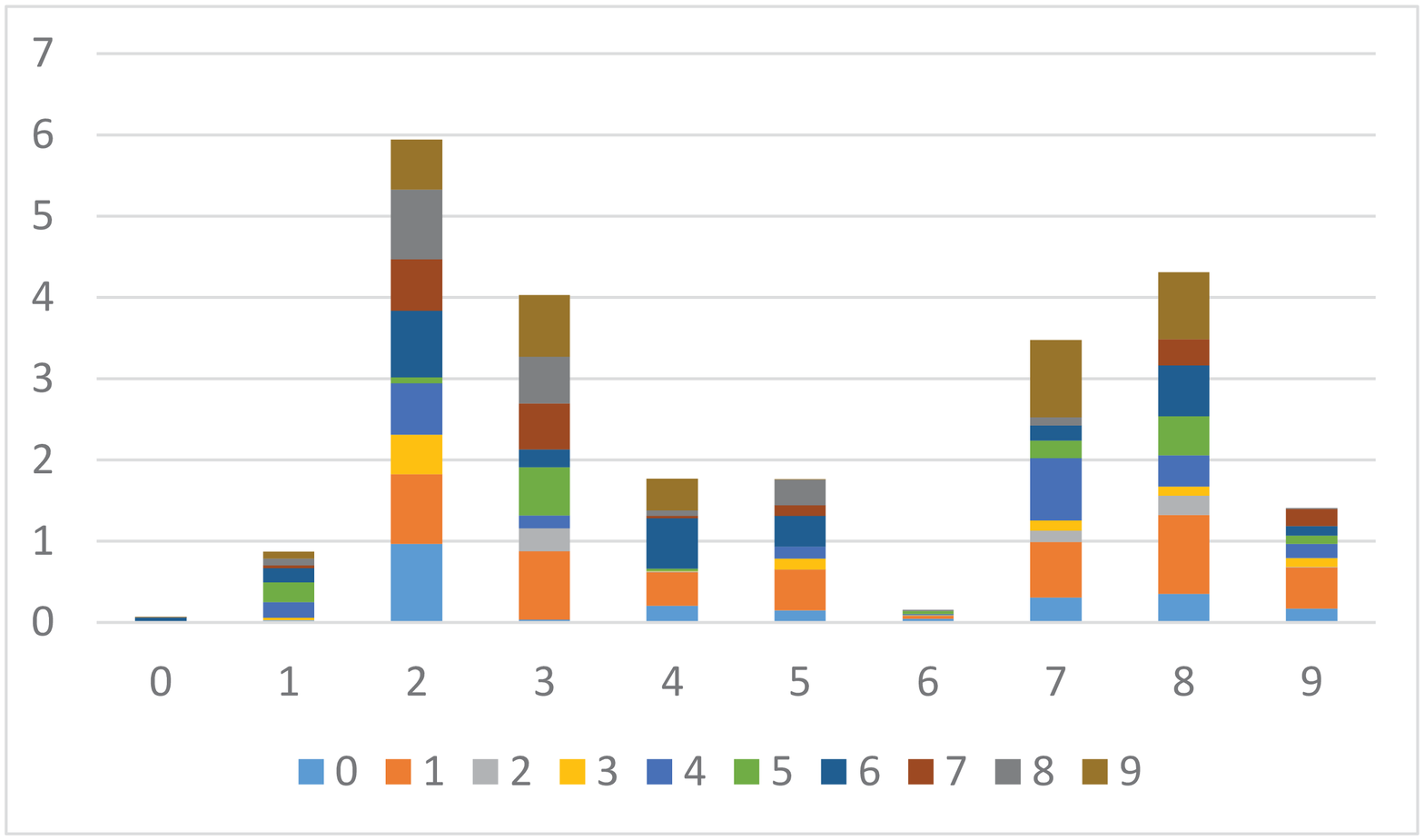}
    \vspace{-0.2cm}
    \caption{\small Hardness of Target}
    \label{figure: 10epotarget}
 \end{minipage}
\vspace{-0.4cm}
\end{figure}

%%Unlike untargeted attack,
{\bf Success Rate (SR).\/} We first categorize the easy and hard cases in multi-step targeted attacks using
%%the statistical measures, such as
the high and low SR or the large or small fraction of adversarial examples misclassified. Table \ref{table:targeted10epoch} shows the results.
For source class digit 1 with 1135 images and attack target digits 0, 2, 6, and 8, the SR is 0.1\%, 85.6\%, 3\%, and 97\% respectively. Clearly, the SRs for misclassifying digit 1 to target digit 0 or 6 (hard cases), are much lower than the SRs for the target digits 8 or 2 (easy cases). Also, the effectiveness of targeted attacks is asymmetrical, i.e., the attack $7\to9$ is much harder with low SR of 0.208 than the reverse attack $9\to7$ with high SR of 0.944.
%%
%%Given these attacks with high and low SRs, it is class-wisely easier to perform some attacks than other attacks. With around 1000 images of input, the SR of some attacks is still so low that they can hardly happen in real world where the input data are limited.
%%}
In Table \ref{table:targeted10epoch}, the last column and the last row are average SR for each source class and each target class respectively.
% It is observed that 1, 6, 9 are the most vulnerable source classes, while 2, 3, 8 are the most easy target classes statistically.
Figure~\ref{figure: 10eposource} shows vulnerable and robust source classes, and Figure \ref{figure: 10epotarget} shows hard and easy target classes using the SR sum. Within each SR bar, different colors indicate different contributions of each digit to build the SR of the attack. It is easy to see that 1, 9, 6 are the top 3 vulnerable source classes while 2, 8, 3 are the top 3 easy target classes.

%% Figures 8+9 have removed. The following text needs to be removed too.
%% Within each bar, different colors indicate different contributions of each digit to the SR of the corresponding attack. It is observed that 1, 6, 9 are the most vulnerable source classes, while 2, 3, 7, 8 are the most easy target classes statistically.
%%
%% Another way of showing

{\bf Degree of Change (DoC).\/} DoC is another good discriminator for easy and hard cases: the higher average DoC means that the attack is harder.
Table \ref{table:degreeofchange1} shows the DoC for target attack on digit 1 and other classes being the target options. Digits 2 and 3 are easy targets with high SR (Table \ref{table:targeted10epoch}) and relatively low DoC. It only takes 5\% or 6.6\% of change in pixels on average to misclassify an image of digit 1 as digit 2 or digit 3 respectively. However, for hard attack $1\to0$, the DoC is 15\% and the SR is 0.1\%.

{\bf Average Entropy.\/} We may also use average entropy to differentiate easy and hard attacks. Two situations have low entropy: benign input whose order of entropy is around $10^{-7}$, and unsuccessful adversarial example. However, there is no linear correlation between high SR and larger entropy for two reasons. (1) When injecting full value of 255 to a pixel at a time, the per-step perturbation may be too large, and such coarse-grained perturbation leads to the change of prediction vector from one-hot source class to one-hot target class directly, which will not increase entropy. (2) The resemblance of source and target images (digits) may play a role. For example, digits 5 and 6, digits 7 and 9 look alike, respectively. And attacks $6\to5$ and $9\to7$ have higher SRs, and are more successful. But their entropy values are smaller than attacks $6 \to 3$ and $9\to8$.

 \begin{table}[ht]
\centering
\centering
\scalebox{0.7}{
\small{
\begin{tabular}{|c|c|c|c|c|c|c|c|c|c|c|c|}
    \hline
    S $\backslash$ T & 0 & 1 & 2 &3&4&5&6&7&8&9& S: avg \\
    \hline
  0 & &0.027 &\textbf{0.970}& 0.039& 0.205 &0.147 &0.049 &0.307& 0.352 &0.170 & 0.252 \\
  \hline
  1 &0.001 & & 0.856& 0.838& 0.415& 0.502& 0.030& 0.686& \textbf{0.970} &0.510& \textbf{0.534}\\
  \hline
  2 &0.001 &0.006& & \textbf{0.285}& 0.007& 0.003 &0.009 &0.136& 0.237 &0.004&0.076 \\
  \hline
   3 &0.001 &0.027 &\text{0.483} & &0.005& 0.136 &0.003 &0.125& 0.114 &0.110 &0.112 \\ \hline
4 &0.000 & 0.188 & 0.633& 0.155& & 0.145 &0.013 &\textbf{0.768}& 0.386& 0.173&0.273 \\
\hline
5& 0.013& 0.246& 0.077& \textbf{0.592} &0.033& & 0.037& 0.217& 0.478& 0.105 &0.120 \\
\hline
6& 0.040& 0.176& \textbf{0.815}& 0.223& 0.618 &0.382& & 0.183 &0.630& 0.116 &0.354 \\ \hline
7& 0.003& 0.034& \textbf{0.636}& 0.562& 0.027& 0.129& 0.000& & 0.320& 0.208 &0.213  \\ \hline
8& 0.003& 0.086& \textbf{0.858}& 0.575& 0.071& 0.317& 0.016& 0.107& & 0.015 &0.228\\ \hline
9& 0.010& 0.084& 0.613& 0.761 &0.387 &0.003 &0.000 &\textbf{0.944}& 0.825& &0.403 \\ \hline
T: avg & 0.008 & 0.097 & \textbf{0.660} & 0.448 & 0.196 & 0.196 & 0.017 & 0.386 & 0.479 & 0.157 &\\ \hline
\end{tabular}
}}
%\vspace{-0.4cm}
\caption{\small SR of adversarial examples in Jacobian-based attack.}
\label{table:targeted10epoch}
%\vspace{-0.4cm}
\end{table}

\vspace{-0.3cm}

 \begin{figure}[ht]
 \centerline{\includegraphics[width=8.7cm]{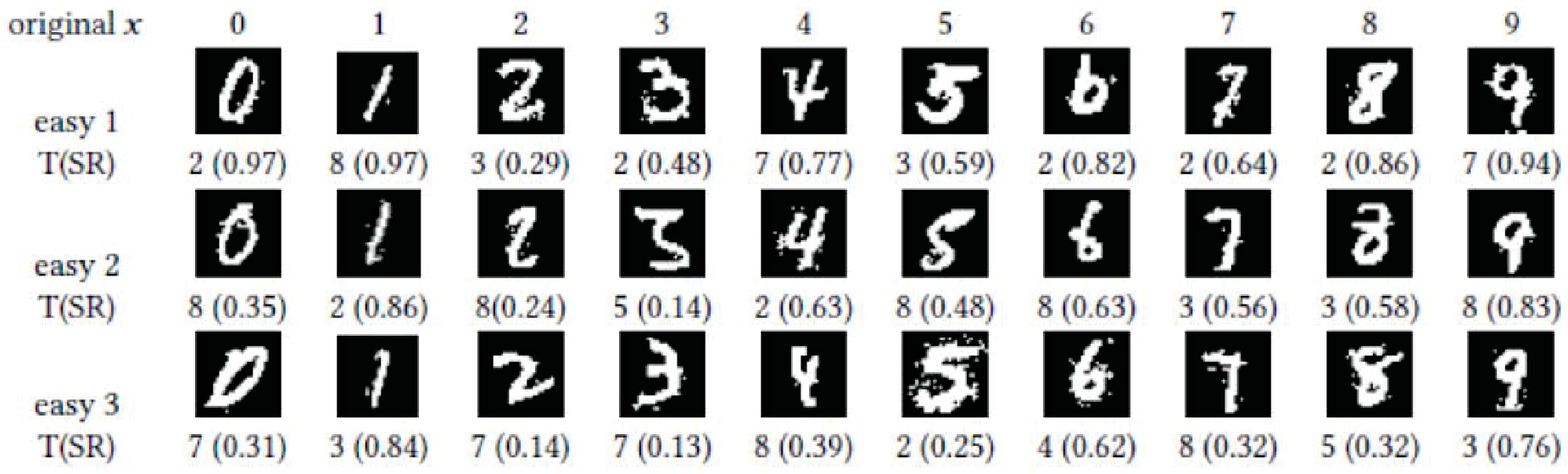}}
\vspace{-0.2cm}
\caption{\small Top 3 easy cases per target in Jacobian-based Attack}
\label{figure:top3targeted}
 \vspace{-0.4cm}
\end{figure}

 %\vspace{-0.3cm}

\begin{table}[ht]
\centering
\scalebox{0.8}{
\small
\begin{tabular}{|c|c|c|c|c|c|c|c|c|c|}\hline
  Target & 0  & 2 & 3 & 4 & 5 & 6 & 7 & 8 & 9\\ \hline
  DoC & 0.150 & 0.050 & 0.066 & 0.101 & 0.102 & 0.148 & 0.066 & 0.029 & 0.093\\ \hline
  Entropy &0.026 &  0.069 & 0.068& 0.03 & 0.064& 0.017& 0.05& 0.067 & 0.048  \\ \hline
\end{tabular}
}
\caption{\small DoC and entropy of 1135 images of digit 1.}
\label{table:degreeofchange1}
 \vspace{-0.4cm}
\end{table}

{\bf Takeaway Remarks.\/} (1) Designing a strong attack requires to trade off between larger per-step perturbation and minimal degree of change. An attack is considered strong if it succeeds with high confidence or if the adversarial prediction vector is one-hot vector. However,
not all successful attacks are accompanied with high SR. Often the successful multi-step attacks may lower the confidence (the probability) of prediction to a greater extent. Moreover, targeted attack is hard to develop and pays higher cost to produce successful adversarial examples. We measure the time for performing untargeted FGSM attack and the time for performing Jacobian based targeted attack using Intel @ Core i5-2300 CPU. The former takes $0.53$ second and the latter takes $3.7$ seconds per instance, which is 7 times on average for the same input.
%% Wenqi, FGSM is untargeted, right???
(2) Even with the same attack algorithm, top three easy cases may vary notably with respect to SR, DoC and entropy, so do the hard cases.
Figure~\ref{figure:top3targeted} shows the top 3 easy Jacobian based attacks. The number in the bracket is the SR. For images of digit 5, the top 1 easy attack is $5 \to 3$ with SR of 59\%, and the top 3 easy attack is $5 \to 2$ with SR of only 25\%, though the attack is successful with only 86 perturbed pixels. Also the crafted image of digit 5 still looks like digit 5 visually, but the perturbation noise is visible too.
These empirical evidences show some strong and complex connection between benign input, adversarial input, loss function, and prediction vector, which inspires us to investigate the effectiveness of adversarial
examples from another set of factors related to adversarial learning and DNN training in Section 6.

%% Wenqi, suggest to remove the following since convergence on two types of attacks are not useful for us at this moment.
%%When comparing the easy attack distributions in Figure \ref{figure:top3targeted} with the easy FGSM attacks in Table~\ref{table:top3untargeted}, we observe some similarity: for source class 1, the top three easy cases under untargeted FGSM are the digits 8, 2, 3 with SRs of 0.524, 0.123, 0.092 respectively, and the top three easy cases under targeted Jacobian are the same three classes in the same order, i.e., digits 8, 2, 3 with SRs of 0.97, 0.86, and 0.84 respectively.
%%
%%: decreasing the loss function on one class will result in an increase in the prediction probability on that class. This observation inspires us to investigate the influencing factors on the behavior of the effectiveness of adversarial
%examples, which is the focus of our empirical analysis in Section 6.}}

\section{Attack Effect of DNN Frameworks}
\label{DNN}
%%on the Effectiveness of Attack Example}

In this section, we characterize the effect of different settings of hyperparameters and different DNN frameworks on the attack effectiveness of adversarial examples.
We choose the number of training epochs to study the impact of overfitting, various sizes of feature maps to compare the effectiveness under different DNN capacity and different DNN frameworks to evaluate their impact on the effectiveness of adversarial attacks.
%% move the remarks to the end.

\subsection{Different Number of Training Epochs}

The first set of experiments reports the presence of inconsistent easy and hard attacks under different training epochs.

%\vspace{-0.2cm}
% \begin{figure*}
% \begin{minipage}{0.24\linewidth}
% \centerline{\includegraphics[width=4.2cm]{figure/1epocheasy.eps}}
%   \vspace{-0.1cm}
%   \subcaption*{\small Top 3 Easy Attacks at 1 epoch}
%   \label{figure: 1epocheasy}
% \end{minipage}
% \begin{minipage}{0.24\linewidth}
% \centerline{\includegraphics[width=4.2cm]{figure/30epocheasy.eps}}
%   \vspace{-0.1cm}
%   \subcaption*{\small Top 3 Easy Attacks at 30 epochs}
%   \label{figure: 30epocheasy}
% \end{minipage}
% \begin{minipage}{0.24\linewidth}
%  \centerline{\includegraphics[width=3.9cm]{figure/epochsource.eps}}
%  \vspace{-0.1cm}
%  \subcaption*{\small Vulnerability of Source (higher SR, more vulnerable)}
%  \label{figure: epochsource}
% \end{minipage}
% \begin{minipage}{0.24\linewidth}
%  \centerline{\includegraphics[width=3.9cm]{figure/epochtarget.eps}}
%   \vspace{-0.1cm}
%   \subcaption*{\small Hardness of Target (lower SR, harder target)}
%   \label{figure: epochtarget}
% \end{minipage}
% \vspace{-0.4cm}
% \caption{\small Jacobian-based targeted attack under different training epochs}
% \label{figure: statiepoch}
%  %\vspace{-0.4cm}
% \end{figure*}

\begin{figure}[ht]
\begin{minipage}{0.49\linewidth}
 \centerline{\includegraphics[width=4.1cm]{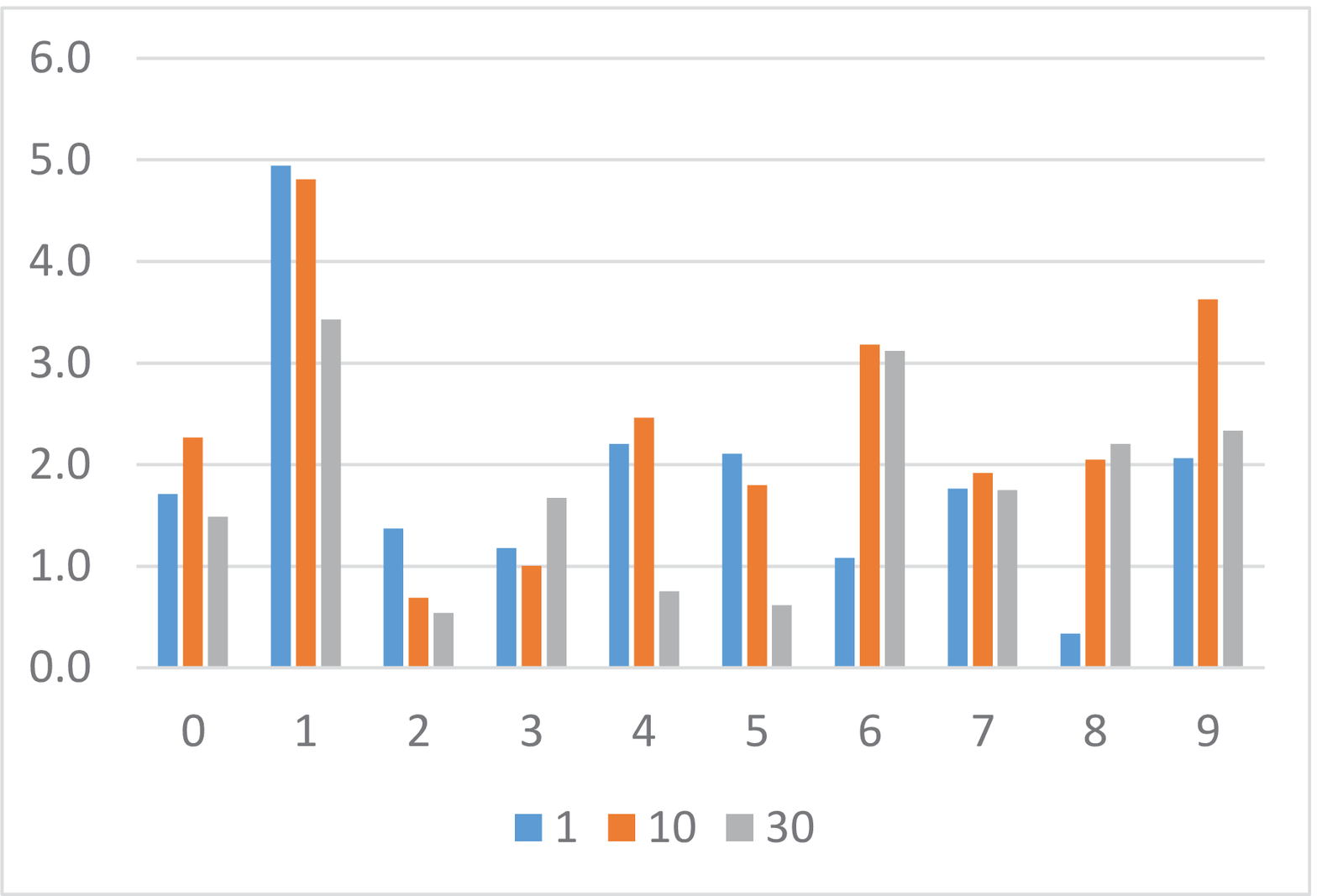}}
 \vspace{-0.1cm}
 \subcaption*{\small Vulnerability of Source}
 \label{figure: epochsource}
\end{minipage}
\begin{minipage}{0.50\linewidth}
 \centerline{\includegraphics[width=4.1cm]{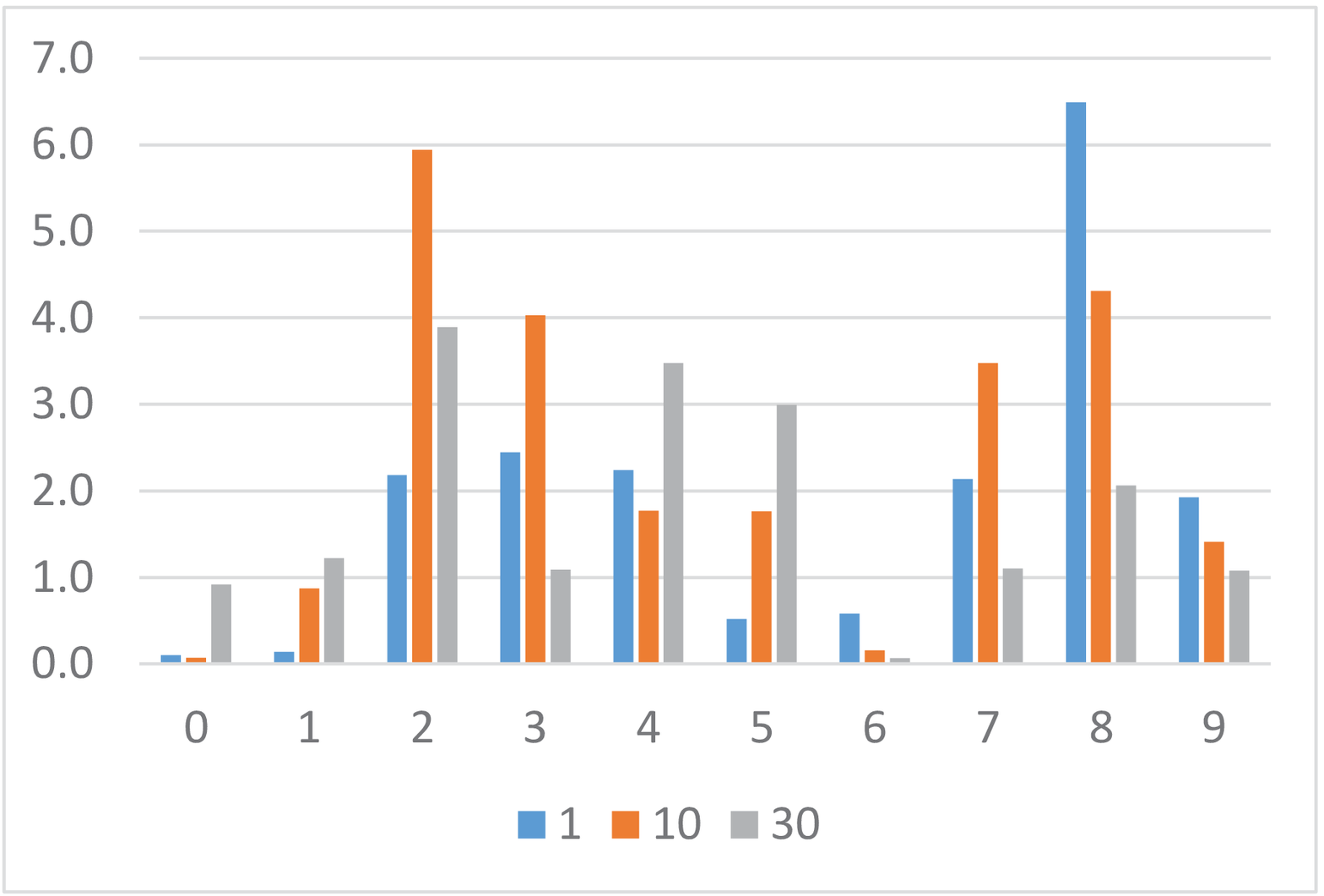}}
  \vspace{-0.1cm}
  \subcaption*{\small Hardness of Target}
  \label{figure: epochtarget}
\end{minipage}
\vspace{-0.4cm}
\caption{\small Impact of training epochs: higher SR, more vulnerable and lower SR, harder attack.}
\label{figure: statiepoch}
 %\vspace{-0.4cm}
\end{figure}
\vspace{-0.1cm}

\begin{figure}[ht]
\centering
\includegraphics[scale=.37]{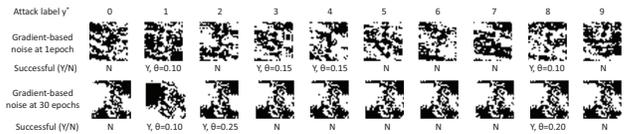}
\vspace{-0.4cm}
\caption{\small Loss Function-Based Noise at 1 and 30 epochs}
\label{figure:epochFGSM}
 \vspace{-0.4cm}
\end{figure}

We study easy and hard cases using Jacobian-based targeted attack with a DNN model trained under three different settings of epochs: 1 epoch (underfitting), 10 epochs (TensorFlow default) and 30 epochs (overfitting). Figure \ref{figure: statiepoch} compares the vulnerability of source class and the hardness of target class. The height of the bar demonstrates the sum of SRs for each of the source classes (left figure) or target classes (right figure). We highlight two interesting observations: (1) Statistically, digit 1 is the most vulnerable source class for all settings of epochs and digit 8 is the most easy attack target for 1 epoch of training. For DNN with 30 epochs of training, digits 1 and 6 are the most vulnerable source classes, and digits 2, 4, and 5 are the most easy targets. Both results indicate different behavior of easy and hard attacks compared to 10-epoch results, where digits 1 and 9 are the most vulnerable source classes, and digits 2, 3, and 8 are the most easy targets.
%%
%% Wenqi, these info are not derived from Fig9, thus should omit.
%% {\color{red}{ In fact, digit 8 is class-wisely the most easy attack for all sources except digit 7. The most easy target for source digit 7 is digit 9.}}
%
%
% \begin{minipage}{0.33\linewidth}
%  \centerline{\includegraphics[width=5.0cm]{figure/epochonehot.eps}}
%   \subcaption{Percentage of One-hot Vector}
%   \label{figure: epochonehot}
% \end{minipage}
%
%
(2) The reason why easy and hard attack cases vary under different training epochs is due to the fact that different training accounts for different trained network parameters, which describe the learned features. The different learned feature is reflected on the gradient of loss function and prediction vectors, and subsequently impacts the effectiveness of adversarial examples. Figure \ref{figure:epochFGSM}  visualizes the gradient of loss function for DNN training under 1 epoch or 30 epochs for FGSM attack and Figure \ref{figure:signoneimage} is for 10 epochs, where successful ones marked by their $\theta$ value.  These empirical evidence shows visible inconsistency across different training epochs regarding success or failure of attack, as well as regarding SR and DoC for successful attacks.

%% {\color{red}{For the sake of demonstration, the gradient of loss function, instead of adversarial saliency map, is visualized in Figure \ref{figure:signoneimage} and Figure \ref{figure:epochFGSM} to shows such different learned features. The difference on the adversarial saliency map is provided in appendix \ref{epoch} for reference. }}

 \vspace{-0.2cm}

%%%this is for single column

\vspace{-0.2cm}

% \begin{figure*}
% \centering
% \includegraphics[scale=.80]{figure/signepoch.eps}
% \caption{\small Visualization of Loss Function-Based Noise Injection at 1 and 30 epochs}
% \label{figure:30epochFGSM}
% \end{figure*}
%%%%%%%%%% this is for two column

\subsection{Different Sizes of Feature Maps}

\begin{figure}[ht]
\begin{minipage}{0.49\linewidth}
 \centerline{\includegraphics[width=4.1cm]{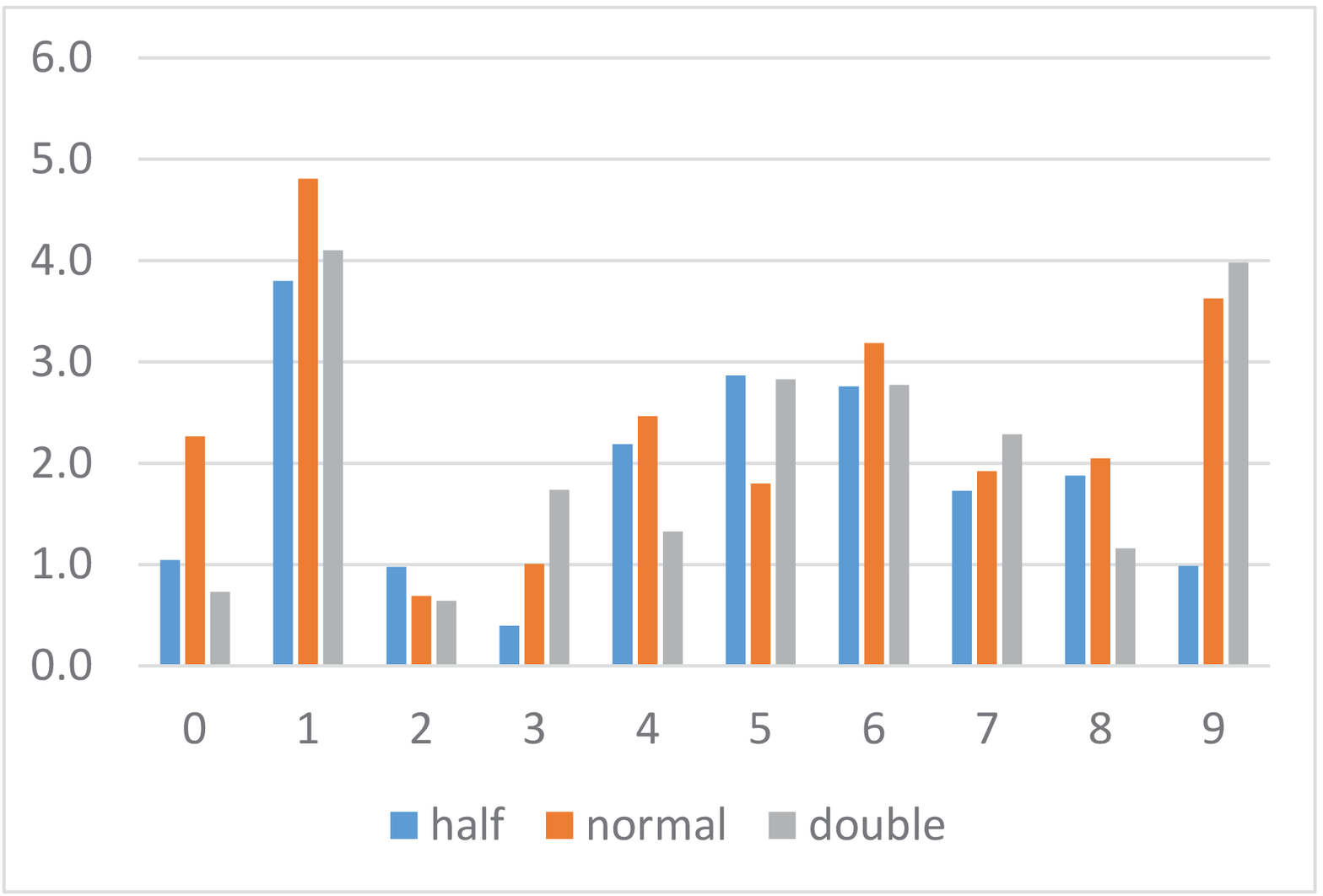}}
 \subcaption*{\small Vulnerability of Source}
 \label{figure: sizesource}
\end{minipage}
\begin{minipage}{0.50\linewidth}
  \centerline{\includegraphics[width=4.1cm]{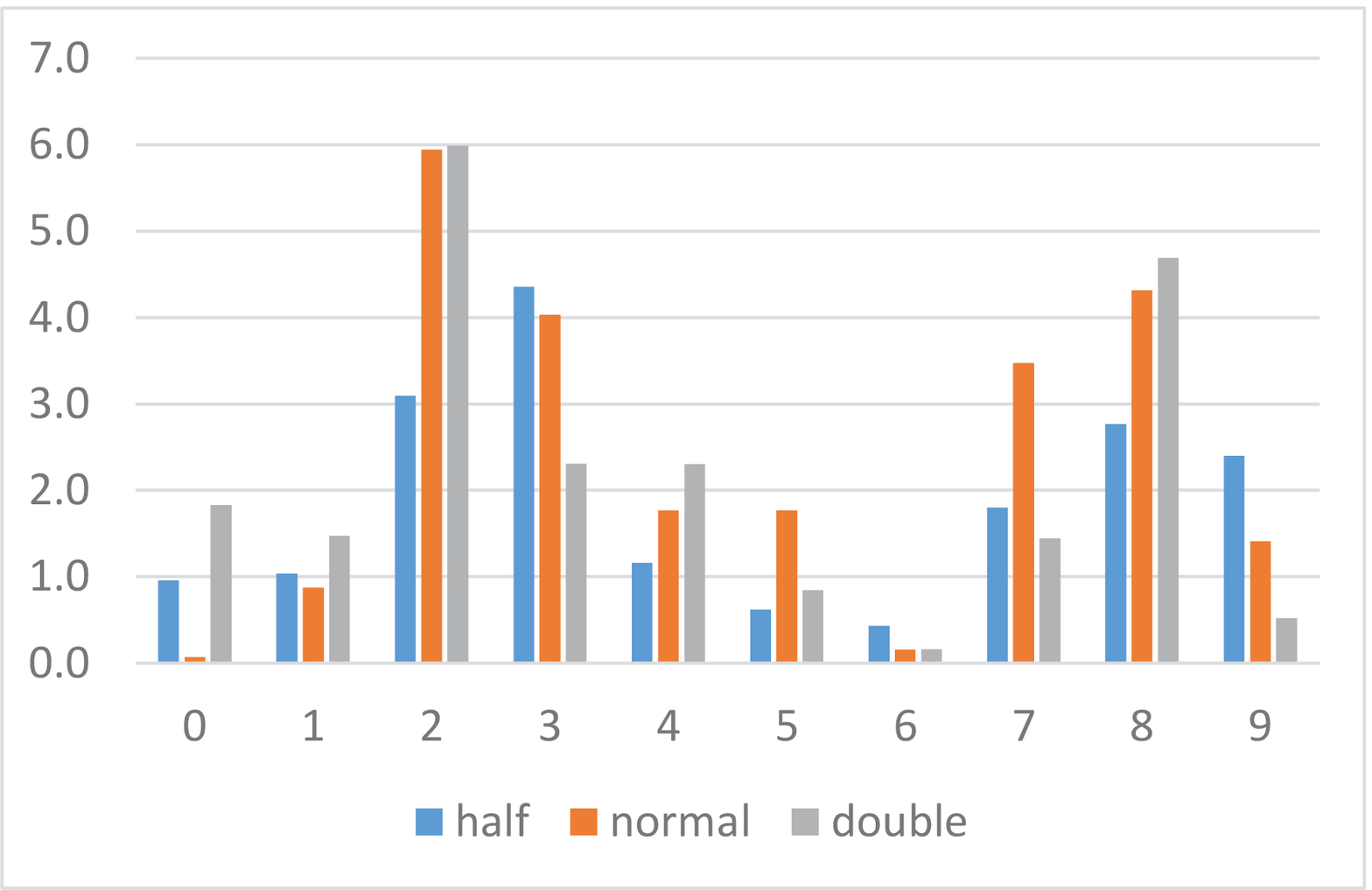}}
  \subcaption*{\small Hardness of Target}
  \label{figure: sizetarget}
\end{minipage}
\vspace{-0.4cm}
\caption{\small Impact of varying sizes of feature maps: higher SR, more vulnerable and lower SR, harder attack.}
\label{figure: statisize}
 \vspace{-0.4cm}
\end{figure}

\begin{figure}[ht]
\centering
\includegraphics[scale=.38]{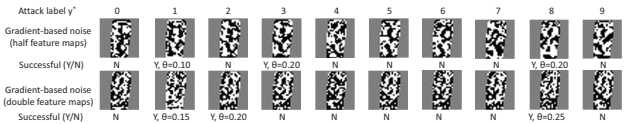}

\vspace{-0.4cm}

\caption{\small Loss function-based noise with different feature maps}
\label{figure:sizeFGSM}

\vspace{-0.4cm}

\end{figure}

%%In this subsection, we
We next study whether different sizes of feature maps have different impacts on the features learned by the DNN model, as the change of learned features will be reflected by the gradients of loss function and adversarial saliency maps, which will impact the behavior of easy and hard attacks. We reduce and double the original number of output features to generate feature map of half and double sizes for the first four DNN layers in TensorFlow. Figure~\ref{figure: statisize} compares three sizes of feature maps: half, original and double on the vulnerability of source classes and the hardness of target classes. For half feature map case, digit 1 is the most vulnerable source class, whereas digits 2 and 3 are the easiest targets. For double feature map case, digit 1 and 9 are the most vulnerable source classes, whereas digits 2 and 8 are the easiest targets. Again, the hard and easy attacks vary for three sizes of feature maps with more easy cases for normal size feature maps. Figure~\ref{figure:sizeFGSM} visualizes the different features learned under half and double feature maps using the gradient of loss function under targeted FGSM attack, and visualization for normal feature map was given in Figure~\ref{figure:signoneimage}. Similar inconsistency is observed across different sizes of feature maps, though the impact of different training epochs on the degree of inconsistency is much larger.

% \begin{figure*}
% \begin{minipage}{0.24\linewidth}
%  \centerline{\includegraphics[width=4.0cm]{figure/halfeasy.eps}}
%   \subcaption*{\small Top 3 Easy Attacks (half)}
%   \label{figure: halfeasy}
% \end{minipage}
% \begin{minipage}{0.24\linewidth}
% \centerline{\includegraphics[width=4.0cm]{figure/doubleeasy.eps}}
%   \subcaption*{\small Top 3 Easy Attacks (double)}
%   \label{figure: doubleeasy}
% \end{minipage}
% \begin{minipage}{0.24\linewidth}
%  \centerline{\includegraphics[width=3.8cm]{figure/sizesource.eps}}
%  \subcaption*{\small Vulnerability of Source (higher SR, more vulnerable)}
%  \label{figure: sizesource}
% \end{minipage}
% \begin{minipage}{0.24\linewidth}
%   \centerline{\includegraphics[width=3.8cm]{figure/sizetarget.eps}}
%   \subcaption*{\small Hardness of Target (lower SR, harder target)}
%   \label{figure: sizetarget}
% \end{minipage}
% \vspace{-0.4cm}
% \caption{\small Jacobian-based targeted attack under different sizes of feature maps}
% \label{figure: statisize}
%  \vspace{-0.4cm}
% \end{figure*}

%% rewrite the following and moved to above
%% {\color{red}{ In addition, we visualize the different features learned by DNN models under different sizes of feature maps using the gradient of loss function in  figure \ref{figure:sizeFGSM}. The comparison with the training epoch indicates that the impact of training epoch is larger than the size of feature map. At least, the gradient of loss function is still around the original digit 1 with various sizes of feature maps while the gradient of loss function is completely different at 1, 10 30 epochs. The visualization on the adversarial saliency map is provided in appendix \ref{size} for reference.}}

\vspace{-0.2cm}

%%%this is for single column

\subsection{Different DNN Frameworks}
%% This subsection can be seen as an extension of framework comparison proposed in \cite{liu2018benchmarking}. While the experiments in previous sections are conducted on TensorFlow, we perform untargeted FGSM attacks on Caffe, Theano and Torch respectively. Note that Torch is in the form of PyTorch in our experiment.

\begin{figure}[ht]
\centering
\includegraphics[scale=.28]{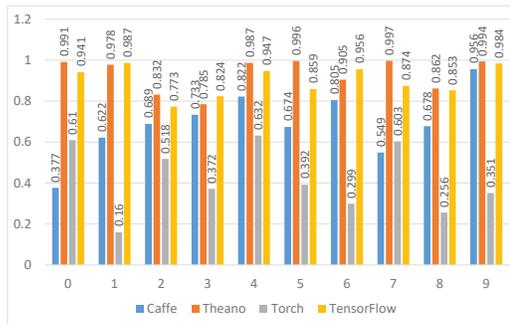}
\vspace{-0.4cm}
\caption{\small SR of untargeted FGSM with different frameworks}
\label{figure:frameworkSR}
 \vspace{-0.4cm}
\end{figure}

\begin{figure}[ht]
\centering
\includegraphics[scale=.38]{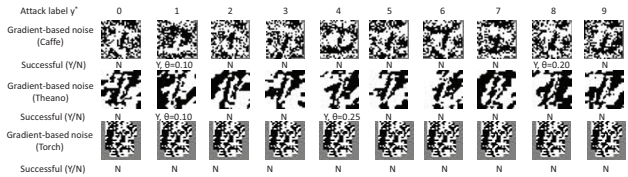}
\vspace{-0.4cm}
\caption{\small Loss function-based noise with frameworks}
\label{figure:frameworkFGSM}
 \vspace{-0.4cm}
\end{figure}

We next evaluate the impact of different DNN frameworks on the effectiveness and divergence of adversarial examples.
Figure \ref{figure:frameworkSR} reports the comparison results for TensorFlow, Caffe,  Theano and Torch. Clearly, TensorFlow and Theano are consistently more vulnerable under FGSM, followed by Caffe and Torch.
%% which are relatively harder.
Figure \ref{figure:frameworkFGSM} visualizes the gradient of loss function based adversarial perturbation using DNN model trained by Caffe, Theano and Torch respectively. It is clear that different frameworks lead to different features learned by their DNN model, which contributes to their different influence on the effectiveness of adversarial attacks with respect to easy and hard cases.
Figure \ref{figure:frameworkFGSM} also exposes some inherent problems in FGSM attack method. The crafting rule in FGSM treats all pixels equally, which may be inefficient since the gradients reflected in the input data for each pixel are not the same. While the positive and negative signs of the sign function are useful, assigning the magnitude of sign function to 1 is not effective in many cases. For example, in Torch, the identical perturbation noise smooths the numerical difference of the gradient of the loss function on different targets. Thus, the attack does not make full use of these gradients. This may contribute to the low SR on untargeted FGSM attacks in Torch.

%% Due to space, we remove the following
%% Second, it is observed that the gradient values toward digit 0 and digit 3 have high degree of similarity (almost identical) for Torch, indicating that the classes of digit 0 and digit 3 are very similar in features learned by Torch at 10 epochs. This is also true for digits 7 and 8 (Table \ref{table:whytorch} in Appendix \ref{framework}). Meanwhile, the gradient of loss function toward target digits 5, 8 9 is more similar than other target digits in Caffe, and digits 4 and 9 in Theano.}} This phenomenon indicates that the learned features for these classes are more similar. If they are grouped together, then the classes within the group are more alike, which could lead to easy transformation between each other, and gives rise to easy attacks. However, such closeness of class labels also leads to situations where a coarse-grained perturbation like changing pixel value to 255 at a time can easily go beyond the distance boundary constraints defined by the attack, especially when the features learned about the two classes are very close. Consequently, hard cases in adversarial attacks are the results of large dissimilarity of the two classes and inappropriate crafting rules.

{\bf Takeaway Remarks.\/}
{\em First}, adversarial attacks heavily rely on the gradient of loss function and prediction vector produced by the trained DNN model, and such gradient is determined by the parameters in the DNN function learned during the training process. Both the hyperparameters, which influence on how the training process will be conducted, and the learned parameters, which are the fixed components of the trained model, will impact the computation of gradient of loss function and prediction vector during the generation of adversarial examples, regardless of specific attack algorithms, and subsequently impact the effectiveness and divergence of adversarial attacks. {\em Second}, for the adversarial examples crafted using the same attack method, easy and hard attacks tend to vary under different hyperparameters and across different DNN frameworks, indicating that the effectiveness of adversarial attacks is inconsistent and unpredictable across different DNN frameworks. Such inconsistency also presents under diverse settings of hyperparameters used for DNN training within the same DNN framework.

%% We not only show the contribution of each label on the destination label for attacks in all three frameworks, but also summarize the overall SR for each label as well. The results show that the effectiveness of adversarial attacks behaves totally different. The detailed experiment results are provided in \emph{Appendix A.4}.
% \begin{figure*}
% \begin{minipage}{0.24\linewidth}
%  \centerline{\includegraphics[width=4.0cm]{figure/FGSMcaffe.eps}}
%  \vspace{-0.1cm}
%  \subcaption*{\small Caffe}
%  \label{figure: caffe}
% \end{minipage}
% \begin{minipage}{0.24\linewidth}
%  \centerline{\includegraphics[width=4.0cm]{figure/FGSMtheano.eps}}
%   \vspace{-0.1cm}
%   \subcaption*{\small Theano}
%   \label{figure: theano}
% \end{minipage}
% \begin{minipage}{0.24\linewidth}
%  \centerline{\includegraphics[width=4.0cm]{figure/FGSMtorch.eps}}
%   \vspace{-0.1cm}
%   \subcaption*{\small Torch}
%   \label{figure: torch}
% \end{minipage}
% \begin{minipage}{0.24\linewidth}
%  \centerline{\includegraphics[width=4.0cm]{figure/framework.eps}}
%   \vspace{-0.1cm}
%   \subcaption*{\Small Comparison of Frameworks (SR)}
%   \label{figure: framework}
% \end{minipage}
% \vspace{-0.4cm}
% \caption{\Small SR of Untargeted FGSM with different frameworks}
% \label{figure:FGSMframework}
%  \vspace{-0.4cm}
% \end{figure*}

% If space is an issue, we show only the comparison of Frameworks (SR). Move the other to appendix.

%\vspace{-0.3cm}

\begin{figure}
\begin{minipage}{0.49\linewidth}
 \centerline{\includegraphics[width=4.0cm]{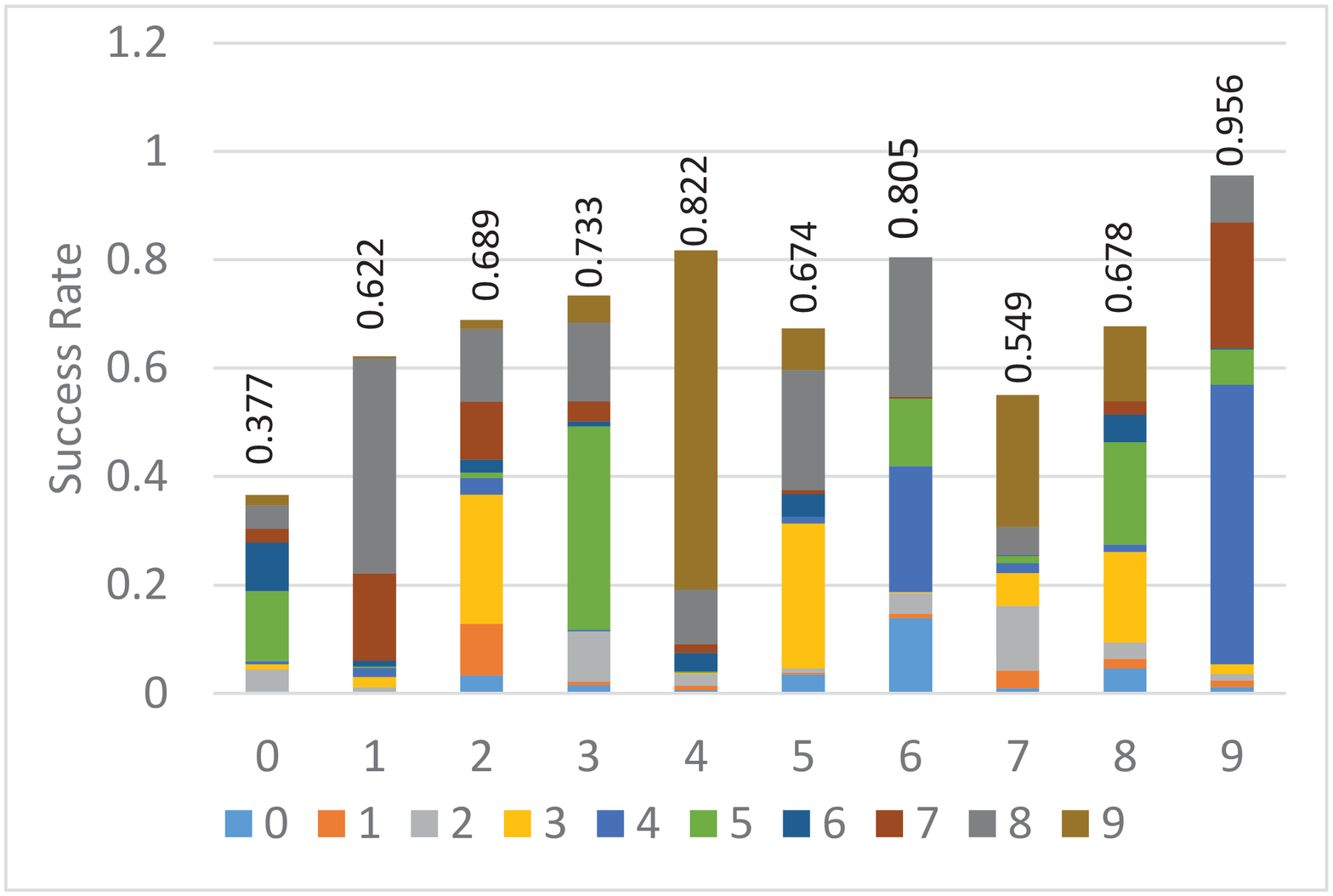}}
 \vspace{-0.1cm}
 \subcaption*{\small Caffe}
 \label{figure: caffe}
\end{minipage}
\begin{minipage}{0.50\linewidth}
 \centerline{\includegraphics[width=4.0cm]{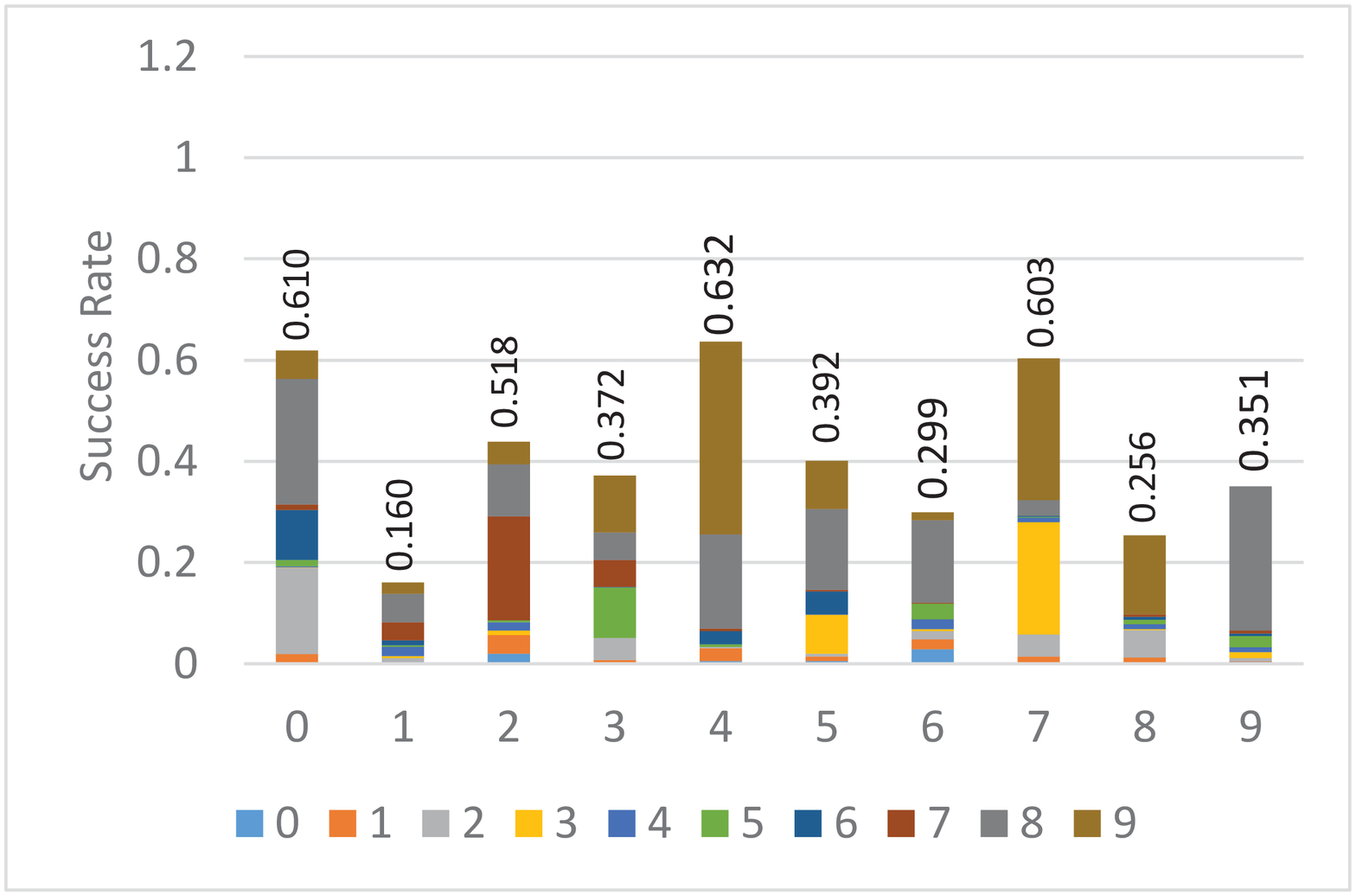}}
  \vspace{-0.1cm}
  \subcaption*{\small Torch}
  \label{figure: torch}
\end{minipage}
\vspace{-0.4cm}
\caption{\Small SR of untargeted FGSM with Caffe and Torch.}
\label{figure:FGSMframework}
 \vspace{-0.4cm}
\end{figure}

Figure \ref{figure:FGSMframework} shows the SR of untargeted FGSM attack on each source class for only Caffe and Torch.
Within each SR bar, different colors indicate different contributions of destination classes to building the attack SR. It is easy to see that the top 3 vulnerable source classes in Caffe are very different from that in Torch.

\section{Attack Mitigation Strategies}
\label{defense}

We have characterized the effectiveness of adversarial examples in deep learning through general formulation, extensive empirical evidences, and systematic study of successful attacks and their divergence in terms of easy and hard cases. Motivated by the results of this study, we propose some attack mitigation strategies from two perspectives: prediction phase mitigation and model training phase mitigation.

{\bf Prediction Phase Mitigation.\/} We have shown that (1) successful attacks (targeted or untargeted) often do not agree on the same noise level ($\theta$) under the same crafting rule $R(\,)$; (2) the same $\theta$ value that generates one successful adversarial example against a benign input $x$ of class $C_x$ may not work effectively for another benign input of the same class; and (3) the destination class of untargeted attacks is not uniformly random. Similarly, some target classes are much harder to attack under the same attack scheme. These observations inspire us to propose two prediction phase mitigation strategies: consensus based mitigation and time-out based mitigation, which are independent of trained DNN model and specific DNN framework used for training.

{\bf Consensus based Mitigation.\/} For each prediction query with an input $x$, the prediction API from the DNN as a service provider will generate $q$ input queries $x_1, \dots, x_q$ such that the prediction result is only accepted by a client when the majority reaches a consensus. Given that all adversarial attacks do not respond to the different input data of the same class consistently, such data-diversity based consensus can be an economical and effective mitigation strategy.
There are several ways to generate such $q$ query inputs. For images, one can leverage computer vision and computer graphics techniques to generate alternatives views of the same image. Also, the consensus protocol can be decentralized to make it more resilient to single point of failure~\cite{kosba2016hawk}. Each client may accept a prediction result upon obtaining $q$ consensus votes from the network.

{\bf Time-out based Mitigation.\/}  For each type of prediction queries, a time-out threshold is pre-set by the DNN as a service provider. If such a query input $x$ is compromised by an adversarial example $x_{adv}$ for multi-step attacks, then the time for turning-around prediction for $x_{adv}$ may exceed the normal histogram statistics for this type of prediction task, and thus turn on an alarm. Such time-out threshold can be learned over time or through training. This mitigation strategy can be especially useful for hard attacks, which requires longer iteration rounds to be successful.

{\bf Training Phase Mitigation.\/} We have shown that the effectiveness of adversarial attacks is inconsistent and unpredictable and easy and hard attacks tend to vary under different hyperparameters and across different DNN frameworks. Also adversarial attacks heavily rely on the gradient of loss function and prediction vector produced by the trained DNN model, and such gradient is determined by the parameters in the DNN function learned during the training process. Thus, we propose three training phase mitigation strategies as proactive countermeasures that can be exercised by the DNN as a service provider.

{\bf Data based Ensemble Training.\/} For each of the prediction classes, an adversarial training in conjunction with data driven ensemble is employed. This enables the training set to include sufficient representations of training data for each class, including those that can strengthen the resilience of prediction queries against adversarial examples in the prediction phase. For instance, by studying the hard cases and easy cases of each source class and each target class, we can generate training examples that make the easy attacks harder and make the harder attacks impossible to succeed.

{\bf Hyperparameter based ensemble training.\/} By utilizing different settings of hyperparameters, such as different number of epochs, we can train alternative models and use these diverse models as a collection of candidate models in the prediction phase. There are several ways to implement the consensus for hyperparameter based ensemble. For instance, for each prediction query with input $x$, a subset of hyperparameter-varied models will be selected to produce prediction results and collect consensus accordingly. Round robin, random, weighted round robin, power of two~\cite{richa2001power} or generalized power of choice~\cite{park2011generalization} can be employed to implement the selection algorithms.

{\bf DNN Framework based ensemble training.\/} We propose to deploy two types of DNN framework based ensemble training. The first approach is to train a DNN ensemble model for prediction using a number of different deep learning frameworks (e.g., TensorFlow, Caffee, Torch) or using different hyperparameters, such as feature maps, within one framework such as TensorFlow. Recall Section 5, we have shown that same adversarial examples have inconsistent effects when using different sizes of feature maps, different DL frameworks from different DNN software providers due to variations in neural network structures and parallel computation libraries used in their implementations~\cite{liu2018benchmarking}. In addition to the ensemble of final trained DNN models, the second alternative approach for DNN framework based ensemble training is to allow multiple DNN models trained over the same training dataset to co-exist for serving the prediction queries.

Both approaches provide a number of advantages. First, different models respond to the same adversarial example very differently in terms of easy and hard attacks as shown in Section 5, thus the prediction API can detect inconsistency, spot the attack attempts, and mitigate risks proactively. Second, one can also integrate the two alternative approaches for the DNN framework based ensemble training, to further strength the attack resilience through combining multi-framework or multi-configuration of hyperparameter based ensemble training with multi-view based prediction ensemble. Such integrated approach can provide a larger pool of alternative trained models for both training and prediction-based consensus, which further strengthens the prediction query based consensus.

Finally, we would like to note that our proposed DNN framework ensemble approaches are different from the cross ML-models based ensemble strategy, which provides ensemble learning model by integrating different machine learning models, such as SVM, Decision Tree, with DNNs, in order to train a prediction model on the same training dataset~\cite{papernot2016transferability}. The recent study of the transferability of adversarial attacks has shown that using the cross ML models based ensemble learning may not be effective under transferability of untargeted adversarial attacks~\cite{tramer2017space,papernot2017practical,papernot2016transferability,liu2016delving}. As pointed out in~\cite{liu2016delving}, the transferability only works under untargeted adversarial attacks, and targeted adversarial attacks do not transfer. Therefore, our proposed two types of multi-framework based ensemble strategies can be viewed as a step forward towards developing a unifying mitigation architecture for both targeted and untargeted adversarial attacks.

\vspace{-0.1cm}
\section{Related Work}

%%The burgeoning success of deep learning has raised the security and privacy concerns as more and more tasks are accompanied with sensitive data \cite{barreno2006can, papernot2016towards}. The existence of adversarial examples has inspired a sizable body of research on

%\cite{madry2017towards, szegedy2013intriguing,kurakin2016adversarial,papernot2016limitations,evtimov2017robust,goodfellow2014explaining,carlini2017towards,sharif2016accessorize,carlini2016hidden,kurakin2016physical,xie2017adversarial,moosavi2016deepfool,elsayed2018adversarial,nguyen2015deep,ahmed2017poster},

Research on adversarial attacks in deep learning can be classified into two broad categories: attack algorithms and defense proposals~\cite{metzen2017detecting,carlini2017detection,goodfellow2014explaining, papernot2016distillation,gu2014towards,cao2017mitigating,goswami2018unravelling,zhao2018retrieval,rakin2018robust,meng2017magnet}.
%% Adversarial attacks can happen both in training phases and prediction phases. Typical training phase attacks put adversarial training data into the original training data to mis-train the network model \cite{madry2017towards}.  As for testing phase attacks, adversaries craft the benign data to adversarial data during validation.  Due to transferability  \cite{tramer2017space,liu2016delving,papernot2016transferability,papernot2017practical}, adversarial examples generated from one deep learning model can be transferred to fool other deep learning models. This indicates that even the adversaries without perfect knowledge of the DNN model could pose a serious threat to the deep learning model. Given that deep learning is complex and is not yet completely understood, there are more hidden spots that can be utilized for generating adversarial examples, even though there are a number of attacks already. In our study, we focus on existing adversarial attacks that occur during the prediction phase.
Given vulnerability of DNN, there have been quite a few attempts in building a robust system against adversarial examples. Two classes of defense mechanisms have been proposed. The first type of defense is to detect adversarial examples so that malicious data can be removed before prediction \cite{metzen2017detecting,carlini2017detection}. \cite{carlini2017detection} surveys ten recent proposals designed for adversarial example detection. They show that all ten detection methods can be defeated by constructing new loss functions, which makes the adversarial example detection of little use. Besides, the defense mechanism of simply distinguishing between clean and adversarial data is not strong enough. It is better to also correctly classify the carefully-injected adversarial examples.
The second type of defense is to increase robustness by modifying the DNN model, aiming to increase the cost of crafting benign samples into misleading ones \cite{goodfellow2014explaining, papernot2016distillation,gu2014towards,cao2017mitigating,goswami2018unravelling,zhao2018retrieval,rakin2018robust,zantedeschi2017efficient,ahmed2017poster}. Representative defense includes adversarial training \cite{goodfellow2014explaining}, autoencoder-based defense \cite{gu2014towards} and defensive distillation \cite{papernot2016distillation}. While adversarial training is computation-inefficient, the latter two mechanisms require some major modifications on DNN architecture. Region-based defense \cite{cao2017mitigating} is a recently proposed defense mechanism. It generates a number of samples around the input data and regards the label of majority of the samples as the predicted label of the input.

For classification of adversarial attacks, \cite{madry2017towards} shows the impact of network architecture on adversarial robustness, claiming that networks with a larger capacity than needed for correctly classifying natural examples could reliably withstand adversarial attacks. However, larger capacity of the network will increase computation overhead significantly. \cite{ma2018dawn} uses local intrinsic dimensionality in layer-wise hidden representations of DNNs to study adversarial subspaces, while \cite{lu2018limitation} points out its limitation. In spite of a growing number of proposed attacks and defenses, there is a lack of statistical and principled characterization of adversarial attacks, which is critical for systematic instrumentation of mitigation strategies and defense methods.

\vspace{-0.1cm}
\section{Conclusion}

We have taken a holistic approach to study the effectiveness and divergence of adversarial examples and attacks in deep learning systems.  We show that by providing a general formulation and establishing basic principle for adversarial attack algorithm design, we are able to define statistical measures and categorize successful attacks into easy and hard cases. These developments enhance our ability to analyze both convergence and divergence of adversarial behavior with respect to easy and hard attacks, in terms of success rate, degree of change, entropy and fraction of successful adversarial attacks, as well as under different hyperparameters and different DNN frameworks. By leveraging the fact that  adversarial attacks exhibit multi-level inconsistency and unpredictability, regardless specific attack algorithms and adversarial perturbation methods, we put forward both prediction phase mitigation strategies and training phase mitigation strategies against present and future adversarial attacks in deep learning.

%Note that in the new ACM style, the Appendices come before the References.

%\input{cfp}

\begin{acks}
% TODO: For the submission, don't include acknowledgments since they would most likely deanonymize you.
This research is partially support by the National Science Foundation under Grants SaTC 1564097, NSF 1547102, and an IBM Faculty Award.
\end{acks}

 % TODO: replace with your brilliant paper!
%\clearpage
%\bibliographystyle{ACM-Reference-Format}

%\bibliography{ccs-sample}

\end{document}